\documentclass[lettersize,journal]{IEEEtran}  
\usepackage{amssymb,amsmath,amsfonts}  
\usepackage{algorithmic,algorithm}
\usepackage{array} 
\usepackage{booktabs} 
\usepackage{tabularx} 
\usepackage{makecell} 
\usepackage{threeparttable} 
\usepackage{multirow}
\usepackage{graphicx} 
\usepackage[caption=false,font=normalsize,labelfont=sf,textfont=sf]{subfig} 
\usepackage[justification=raggedright,singlelinecheck=false]{caption} 
\usepackage{textcomp} 
\usepackage{verbatim} 
\usepackage[normalem]{ulem} 
\usepackage{stfloats} 
\usepackage{url} 
\usepackage{cite} 
\usepackage{float} 
\usepackage{orcidlink}
\usepackage{hyperref}
\usepackage[table]{xcolor}
\usepackage[normalem]{ulem}

\begin{document}

\title{Tackling Incomplete Data in Air Quality Prediction: A Bayesian Deep Learning Framework for Uncertainty Quantification}

\author{Yuzhuang Pian\orcidlink{0000-0001-6201-7762}, Taiyu Wang\orcidlink{0009-0001-0717-1158}, Shiqi Zhang\orcidlink{0000-0003-1366-8045},~\IEEEmembership{Graduate Student Member, IEEE}, Rui Xu\orcidlink{0000-0002-2090-4408}, and Yonghong Liu\orcidlink{0009-0006-6665-106X}

\thanks{Yuzhuang Pian, Shiqi Zhang, Rui Xu, and Yonghong Liu are with the School of Intelligent Systems Engineering, Sun Yat-sen University, Shenzhen 518107, China; the Guangdong Provincial Key Laboratory of Intelligent Transportation System, Guangzhou 510006, China; and the Guangdong Provincial Engineering Research Center for Traffic Environmental Monitoring and Control, Guangzhou 510006, China((email: \{pianyzh, zhangshq73, xurui27, liuyh3\}@mail2.sysu.edu.cn).
Taiyu Wang is with the School of Intelligent Systems Engineering, Sun Yat-sen University, Shenzhen 518107, China(email:wangty56@mail3.sysu.edu.cn).
}

\thanks{Corresponding authors: Yonghong Liu (email: liuyh3@mail.sysu.edu.cn)}

\thanks{Manuscript received 0, 2025; revised 0, 2025.}}

\markboth{Journal of \LaTeX\ Class Files,~Vol.~14, No.~8, April~2025}%
{Shell \MakeLowercase{\textit{et al.}}: A Sample Article Using IEEEtran.cls for IEEE Journals}

\maketitle

\begin{abstract}
Accurate air quality forecasts are vital for public health alerts, exposure assessment, and emissions control. In practice, observational data are often missing in varying proportions and patterns due to collection and transmission issues. These incomplete spatio-temporal records-combined with the lack of explicit mechanisms for modeling measurement noise and predictive uncertainty—impede reliable inference and risk assessment and can lead to overconfident extrapolation. To address these challenges, we propose an end-to-end framework, the channel gated learning unit based spatio-temporal bayesian neural field (CGLU-BNF). It uses Fourier features with a graph attention encoder to capture multiscale spatial dependencies and seasonal temporal dynamics. A channel gated learning unit, equipped with learnable activations and gated residual connections, adaptively filters and amplifies informative features. Bayesian inference jointly optimizes predictive distributions and parameter uncertainty, producing point estimates and calibrated prediction intervals. We conduct a systematic evaluation on two real world datasets, covering four typical missing data patterns and comparing against five state-of-the-art baselines. CGLU-BNF achieves superior prediction accuracy and sharper confidence intervals. In addition, we further validate robustness across multiple prediction horizons and analysis the contribution of extraneous variables. This research lays a foundation for reliable deep learning based spatio-temporal forecasting with incomplete observations in emerging sensing paradigms, such as real world vehicle borne mobile monitoring.
\end{abstract}

\begin{IEEEkeywords}
	Air quality prediction, incomplete data, uncertainty quantification, Bayesian deep learning, CGLU-BNF.
\end{IEEEkeywords}

\section{Introduction} 
\label{Introduction}
\IEEEPARstart{A}{ir} pollution events will cause serious environmental disasters (greenhouse effect\cite{schneider1989greenhouse}, photochemical smog\cite{dickerson1997impact}) and will also lead to an increase in public health risks such as respiratory and cardiovascular diseases\cite{bennitt2025global}, especially particulate matter. Therefore, leveraging sensor observations for air quality prediction is crucial to accurately assess atmospheric conditions and to enable early warnings of pollution events.

\begin{figure}
	\centering
	\includegraphics[width=0.9\columnwidth]{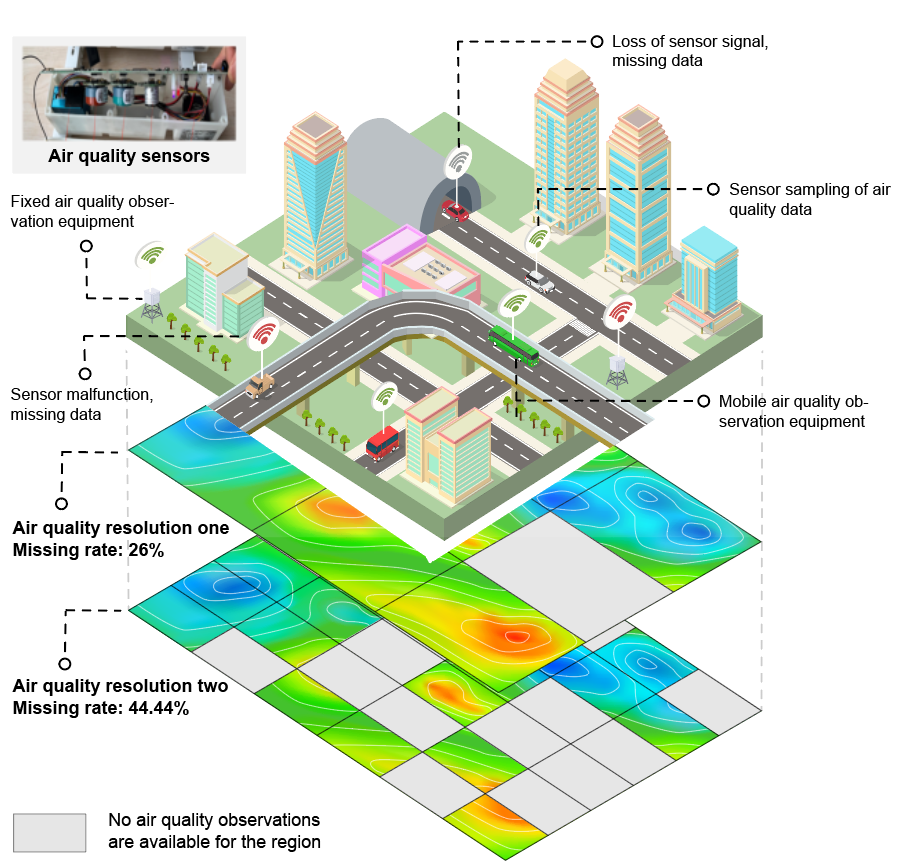}
	\caption{Schematic of air quality observations and missing data patterns. Measurements are collected in real time by fixed monitors and mobile platforms. Two spatial–temporal resolutions are considered: (i) 500 m $\times$ 500 m at 1-h intervals and (ii) 250 m $\times$ 250 m at 1-h intervals.}
	\label{fig1}
\end{figure}

Due to financial constraints and deployment complexities, achieving uniform sensor coverage across urban areas remains challenging, resulting in substantial spatial and temporal gaps in observational data\cite{apte2024high} (Fig.\ref{fig1}). Moreover, sensor malfunctions, maintenance activities, and unstable data transmission further exacerbate data loss\cite{ferrer2021graph}. For instance, at a spatial resolution of 500 m $\times$ 500 m and a temporal resolution of 1 hour, the missing rate can reach approximately 26\%, and when the temporal resolution is refined to 5 minutes, it may soar to 95\%.

\begin{figure}
	\centering
	\includegraphics[width=0.9\columnwidth]{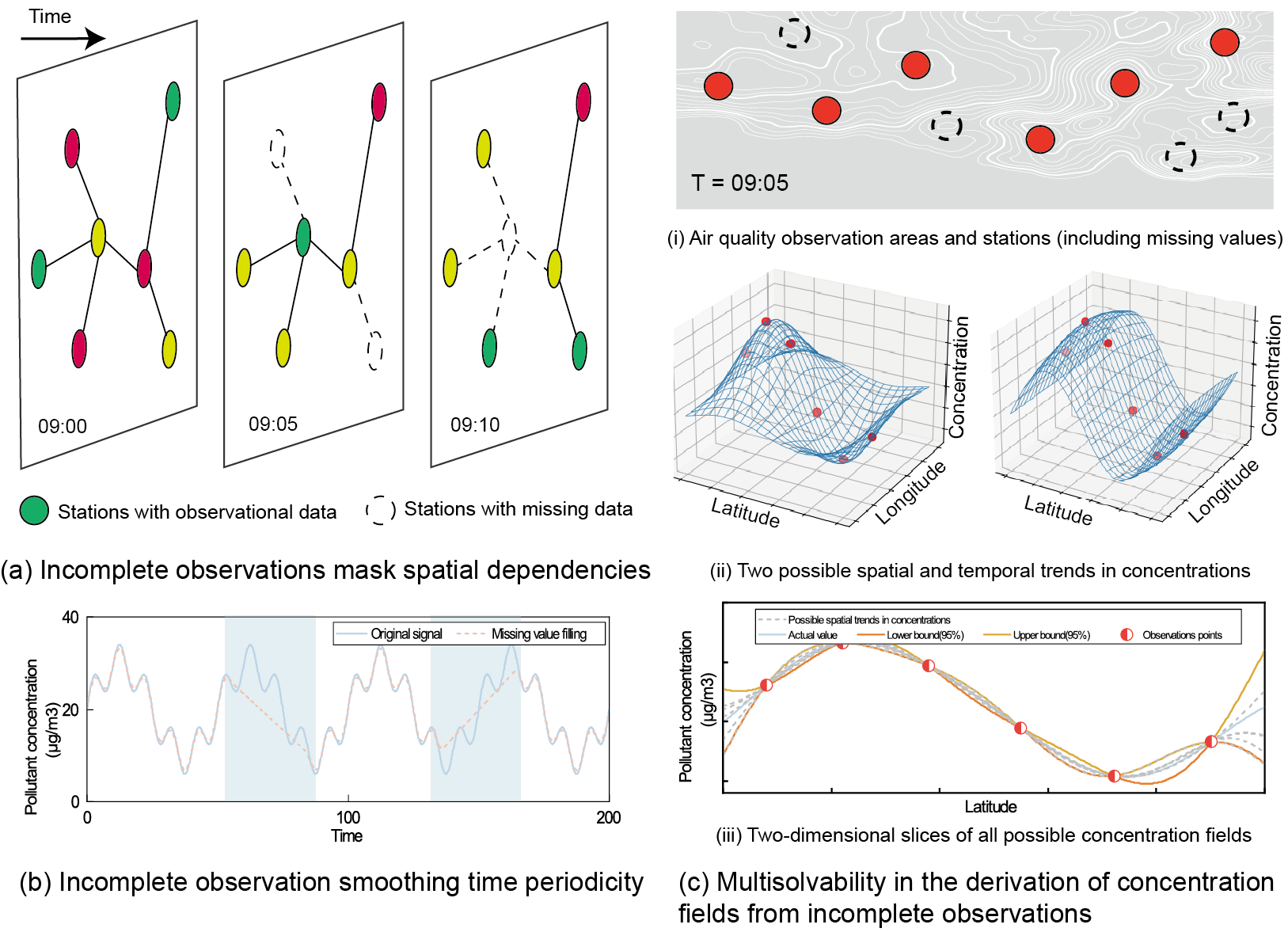}
	\caption{Challenges of air quality prediction tasks with incomplete observations.}
	\label{fig2}
\end{figure}

Incomplete spatio-temporal observations disrupt cross site information flow and distort the covariance structure, obscuring spatial dependence and heterogeneity\cite{xue2024sparse} (Fig.\ref{fig2}(a)). Temporal gaps smooth fluctuations, attenuate periodic and high frequency components, bias the delineation of seasonal cycles, and reduce forecasting accuracy\cite{alsalehy2025improving} (Fig.\ref{fig2}(b)). These effects pose substantial challenges for modeling spatio-temporal dynamics.

Furthermore, under incomplete observations, reconstructing a spatio-temporal field is non-unique: finite measurements with missing entries typically admit a set of feasible solutions rather than a single one. Fig.\ref{fig2}(c) illustrates this with ten stations, four of which lack data. Subject to physical smoothness and statistical consistency, multiple concentration fields can fit the observations equally well (Fig.\ref{fig2}(ii)); their two-dimensional slices expose the resulting spatial multi solution behavior (Fig.\ref{fig2}(iii)). Yet most existing methods return only a single deterministic estimate, neglecting epistemic and aleatoric uncertainties and offering no rigorous assessment of predictive reliability\cite{papadopoulos2024guaranteed}. Therefore, it is essential to develop effective methods for automatically extracting and analyzing meaningful patterns from incomplete data, in order to improve both prediction accuracy and uncertainty estimation.

\begin{figure}
	\centering
	\includegraphics[width=0.9\columnwidth]{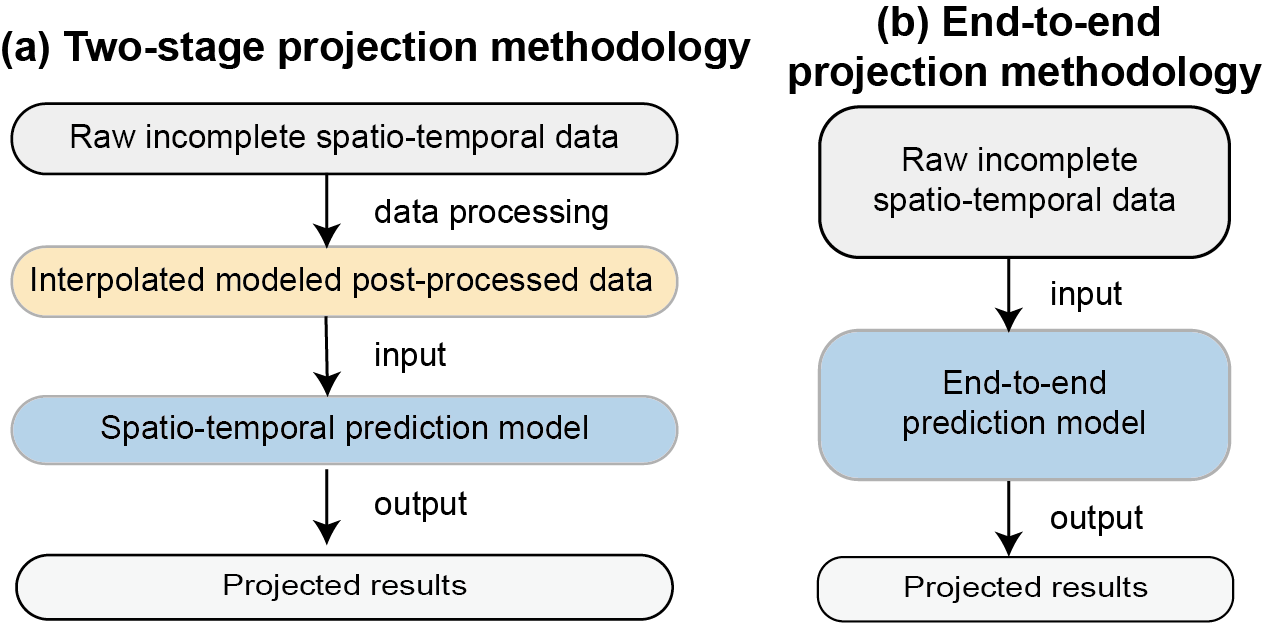}
	\caption{Two methodological ideas for prediction tasks facing incomplete data.}
	\label{fig3}
\end{figure}

Although methods such as ConvLSTM, Transformer, and their variants excel when observations are complete or nearly so, they face limitations with incomplete spatio-temporal data. These models assume that inputs fully and accurately capture the underlying physical state. They also require a fixed spatial grid and uniform time intervals\cite{Wang_2020_CVPR}. Discarding any record that contains missing values causes information loss and introduces estimation bias\cite{little2019statistical}. The two-stage framework\cite{guo2023novel,zhao2024prediction,asaei2023air} was used to address this issue. It first complements missing data and then trains a prediction model on the completed dataset (Fig.\ref{fig3}(a)). However, this direct coupling has limitations. Existing imputation methods struggle to learn the fine grain spatio-temporal features required for high precision air quality forecasting\cite{wang2023traffic}. Moreover, systematic studies on how imputation accuracy affects prediction performance remain scarce.

To address these limitations, end-to-end models have emerged. They optimize feature extraction, data completion, and target prediction simultaneously within a single framework (Fig.\ref{fig3}(b)). Hybrid probabilistic deep models\cite{hamelijnck2021spatio,das2023textitless} are now prominent end-to-end approaches that couple the expressive power of deep networks with the closed-form interpolation and uncertainty quantification of probabilistic methods. In this study, we adopt Gaussian processes (GPs)\cite{sampson1992nonparametric}, treating missing observations as latent variables so that arbitrary missing patterns can be handled within a unified probabilistic framework. However, applying this approach to air quality forecasting presents two major challenges. First, posterior inference incurs substantial computational cost $O({{N}^{3}})$\cite{zhang2023efficient}. Second, selecting key parameters—such as the covariance kernel and mean function—depends heavily on expert domain knowledge\cite{liu2020gaussian}. Thus, designing a probabilistic prediction model that ensures high accuracy and reliable uncertainty quantification while flexibly handling varying degrees of missing data remains a major research challenge.

To overcome these challenges and close existing research gaps, we propose a novel Bayesian deep learning framework: the channel gated learning unit based spatio-temporal bayesian neural field (CGLU-BNF). This framework supports air quality prediction and uncertainty quantification under various missing data patterns. Our study emphasizes improving both prediction accuracy and confidence intervals sharpness when historical spatio-temporal data are incomplete. In the feature extraction, the model first applies a graph attention network to capture spatial dependencies among monitoring stations. It then augments these representations with temporal seasonal and spatial Fourier features to enrich spatio-temporal embeddings in sparse observation scenarios. The channel gated learning unit integrates a learnable activation function, channel attention, and a gated residual mechanism. It non-linearly transforms the input, recalibrates channels, and fuses original and transformed features to adaptively filter and enhance information. Finally, at the Bayesian inference layer, the model uses maximum a posteriori estimation and multi-particle integration to jointly optimize the predictive distribution and parameter uncertainty. The framework directly outputs prediction means along with their confidence intervals. The main contributions are summarized below:

\begin{itemize}
	\item We propose CGLU-BNF, an end-to-end Bayesian deep learning framework that unifies feature extraction, target prediction, and uncertainty quantification. It enables direct forecasting under diverse missing data patterns, eliminating pre-interpolation and other auxiliary imputation steps.
	\item We combine graph attention, temporal harmonics, and spatial Fourier embeddings to build a multilevel spatio-temporal encoder that robustly captures cross scale dependencies and correlations under irregular sampling.
	\item We introduce a channel gated learning unit that integrates channel attention, gated residual networks, and learnable activation functions to dynamically filter and enhance key informative channels, suppress noise, and stabilize training.
	\item Across two real world air quality datasets, CGLU-BNF delivers lower errors and narrower confidence intervals under four typical missing data patterns and across multiple forecast time domains.
\end{itemize}

The remainder of this paper is organized as follows. Section \ref{Related work} reviews related work on air quality forecasting in complete data. Section \ref{Problem formulation} formalizes the problem. Section \ref{methodology} presents the CGLU-BNF framework and details its modules. Section \ref{Experiments} reports experiments under different missing data patterns and rates. Section \ref{discussion} examines the effects of forecast time domains and architectural choices. Section \ref{conclusion} concludes the paper.

\section{Related Work}
\label{Related work}
This section reviews strategies for handling incomplete observations, including explicit imputation in two-stage pipelines and end-to-end predictive approaches.

\subsection{Two-stage Predictive Model}
To address missing data, existing prediction methods can be categorized into two training paradigms: two-stage and end-to-end approaches. In the two-stage framework, missing values are first imputed using statistical or machine learning techniques, after which a prediction model is trained on the completed dataset. Among these, generative approaches—such as variational auto encoders (VAEs) and generative adversarial networks (GANs)—have gained popularity and demonstrate superior interpolation performance. Zhao\cite{zhao2024prediction} combined a Transformer with a GAN: the Transformer extracts temporal features, and the GAN improves data generation and generalization. Asaei\cite{asaei2023air} proposed DAerosol.GAN.NTM, which first imputes missing air quality records with a GAN and then applies a neural turing machine for time series prediction, markedly increasing multi-pollutant forecast accuracy.

Despite their success in interpolation and forecasting, cascaded two-stage frameworks often lack end-to-end synergy because the modules are trained independently. Specifically, existing interpolation methods fail to capture fine grain spatio-temporal dependencies, impairing high precision forecasts and propagating errors downstream\cite{wang2023traffic}. Multi-city studies confirm that pre-prediction interpolation yields suboptimal results, with performance declining sharply as missing rates rise\cite{hua2024impact}. Moreover, most interpolation algorithms produce only single point estimates, preventing reliable uncertainty propagation to the prediction stage\cite{dai2023multiple,chen2019bayesian}.

\subsection{End-to-end Predictive Model}
To mitigate error propagation between task modules, end-to-end predictive framework jointly model the missing data mechanism and the prediction target, enabling imputation and forecasting to share a single objective function. The most direct implementation is the mask-driven deterministic deep network (e.g., STSM\cite{su2024spatial}, HD-TTS\cite{marisca2024graph}), which concatenates observations with a binary mask and uses gating or self-attention to ignore missing entries. Although lightweight, these models struggle to capture long range dependencies under high missing rates or extended gaps\cite{du2023saits}, and they yield only point forecasts without uncertainty estimates\cite{wu2021quantifying}. Generative latent variable approaches address these drawbacks. Variants based on VAEs\cite{rodrigo2024physics} and diffusion models\cite{fan2025diffusion} learn the joint data distribution, sample plausible imputations in latent space, and produce predictions with associated confidence. However, they require dual networks or adversarial training, leading to unstable convergence and high computational cost\cite{wiatrak2019stabilizing}. Consequently, most air quality applications remain limited to small scale or offline settings, falling short of the reliability demanded by multiple missingness model and complex dynamic environments.

To address these limitations, probabilistic approaches based on Gaussian processes  have been extensively explored. They model pollutant concentrations as a continuous spatio-temporal random field and treat missing observations as latent variables inferred jointly in the posterior. They naturally accommodate arbitrary missingness and provide interpolation, prediction, and uncertainty quantification in closed form. However, these methods face key bottlenecks: posterior inference is computationally expensive; performance is sensitive to hyperparameter choices that often rely on domain expertise; and standard kernels are insufficiently flexible for non-smooth dynamics and high dimensional structure\cite{saad2024scalable}. These limitations have motivated hybrid models that couple GP priors with deep spatio-temporal encoders to balance expressiveness and tractability. Hamelijnck et al. presented ST-SVGP\cite{hamelijnck2021spatio}, which integrates a state space formulation with natural gradient variational inference and employs sparse inducing points to model large, incomplete datasets efficiently under non-conjugate likelihoods.

Inspired by previous research, this study presents CGLU-BNF based on GPs, a Bayesian deep learning framework for air quality prediction with incomplete data. The model first fuses a graph attention network with Fourier transforms to extract spatio-temporal features from sparse observations. It then employs a channel gated learning unit to adaptively filter and amplify salient information. Finally, a multi-particle maximum a posteriori ensemble produces predictions and confidence intervals, enabling accurate prediction and uncertainty quantification under multiple mode missing data conditions.

\begin{table*}[t!]
	\centering
	\caption{Comparison of existing air quality prediction methods under incomplete observation conditions.}
	\label{tab1}
	
	\begin{threeparttable}
	\resizebox{\linewidth}{!}{
	\begin{tabular}{llcccl}
		\toprule
		\textbf{\textbf{Paradigm} }               & \textbf{Representative work} & \textbf{ {\makecell[c]{Missing data \\ robustness}}} & \textbf{ {\makecell[c]{Uncertainty \\ quantification}}}  & \textbf{ {\makecell[c]{Computational \\ efficiency}}}   & \textbf{Key limitation}      \\ \midrule
		Complete data forecasting        & CMAQ, ARIMA, STHTNN, etc.                          & ×                                              & ×                                          & 	$\triangle$                            & {\makecell[l]{Requires complete input data}}          \\
		{\makecell[l]{Discriminative methods}} & Kriging, MissForest, MICE, etc.                    & 	$\triangle$                                                 & ×                                          & $\checkmark$                            & Error accumulation                                     \\
		{\makecell[l]{Generative methods}} & {\makecell[l]{DAerosol.GAN.NTM, \\ Transformer-GAN, etc.}}           & $\triangle$                                                 & $\triangle$                                            & ×                               & {\makecell[l]{GAN instability, High computational cost}}                        \\
		{\makecell[l]{Mask-driver models}}  & HD-TTS, STSM, etc.                                & $\triangle$                                                 & ×                                          & $\checkmark$                            & {\makecell[l]{Long gaps, Deteriorate predictions}}                           \\
		{\makecell[l]{Latent variable models}}              & PVGAE, iMMAir, etc.                               &$\checkmark$                                             & $\triangle$                                             & ×                               & {\makecell[l]{Slow convergence, Training expensive}} \\
		{\makecell[l]{Probabilistic predict models}} & ST-SVGP, VSMTGP, etc.                             & $\checkmark$                                              & $\checkmark$                                         & $\triangle$                                & {\makecell[l]{High computational complexity, \\ Limited non-stationary feature representation}}              \\
		{\makecell[l]{Probabilistic predict models}}                                           & \textbf{CGLU-BNF(Our)}                                         & \textbf{$\checkmark$ $\checkmark$}                                            & \textbf{$\checkmark$ $\checkmark$ }                                         & \textbf{$\checkmark$}                           & \textbf{- }                                              \\ 
		\bottomrule
	\end{tabular}
}

\begin{tablenotes}    
	\footnotesize               
	\item[1] Missing data robustness: $\checkmark$ $\checkmark$ (multiple pattern \& long gap), $\checkmark$ (High), $\triangle$ (Moderate), × (Low). 
	\item[2] Uncertainty quantification: $\checkmark$$\checkmark$ (Small interval sharpness), $\checkmark$ (Bayesian), $\triangle$ (Partial or heuristic), × (None). 
	\item[3] Computational efficiency: $\checkmark$ (High), $\triangle$ (Moderate), × (Low)        
\end{tablenotes}            

\end{threeparttable}
\end{table*}

\section{Problem Formulation}
\label{Problem formulation}
\subsection{Incomplete Air Quality Monitoring Information}
Let the air quality monitoring network have nodes $\mathcal{V}=\{1,\dots,N\}$ observed at discrete times $\mathcal{T}=\{{{t}_{1}},\cdots , {{t}_{T}},\cdots, {{t}_{T+H}}\}$. Each station $v\in\mathcal{V}$ has geographic coordinates ${{\mathbf{s}}_{v}}\in {{\mathbb{R}}^{{d}_{s}}}$ (typically longitude and latitude, ${{d}_{s}}=2$). Let ${{y}_{t,v}}$ denote the pollutant concentration at node $v$ and time $t$, and let ${{\mathbf{z}}_{t,v}}$ denote a vector of exogenous covariates (e.g., meteorology, land use). We partition the timeline into a history window ${{\mathcal{T}}_{1}}={{\mathcal{T}}_{\le {{t}_{T}}}}=\{{{t}_{1}},\cdots ,{{t}_{T}}\}$ and a prediction horizon ${{\mathcal{T}}_{2}} = {{\mathcal{T}}_{>{{t}_{T}}}}=\{{{t}_{T+1}},\cdots ,{{t}_{T+H}}\}$. In practice, sensor failures, maintenance, and unstable data transmission cause random or structured missing observations, leading to spatio-temporal discontinuities.

\subsection{Monitoring Graph}
We encode the geospatial topology of the urban monitoring network as a graph $\mathcal{G}=(\mathcal{V},\mathcal{E})$ to capture spatial correlations and enhance prediction under sparse observations. Edges $\mathcal{E}$ represent site connectivity. The adjacency matrix $\mathbf{A}$ is constructed from pairwise distances ${{d}_{ij}}={{\left\| {{\mathbf{s}}_{i}}-{{\mathbf{s}}_{j}} \right\|}_{2}}$, where $\sigma_d$ controls the spatial decay scale.

\begin{equation}
	\label{eq:adjacency matrix}
	{{\mathbf{A}}_{ij}}=\exp (-d_{ij}^{2}/{{\sigma }_{d}}^{2}),i\ne j,{{A}_{ii}}=0
\end{equation}

\subsection{Air Quality Forecasting Under Missing Observations}
Let $\mathcal{O}\subseteq \mathcal{T} \times \mathcal{V}$ be the set of observed indices used in the loss, and let $\eta=|\mathcal{O}|$. Our goal is to learn a stochastic mapping that outputs the predictive mean $\boldsymbol{\mu}_\theta$ and a calibrated uncertainty estimate for $t\in {{\mathcal{T}}_{2}}$.

Formally, we aim to learn a probabilistic model $\rho_{\theta}$ that specifies the conditional distribution of future pollutant concentrations given the available information.


\begin{equation}
	\label{eq:mapping function}
	\rho_{\theta}\big(
	\mathbf{Y}_{\mathcal{T}_2,\mathcal{V}}
	\mid
	\mathbf{Y}_{\mathcal{T}_1,\mathcal{V}},
	\mathbf{Z}_{\mathcal{T}_1,\mathcal{V}},
	\mathcal{T}_{1},
	\mathbf{S}_{\mathcal{V}},
	\mathcal{G}
	\big)
\end{equation}

The model outputs the predictive distribution ${\rho}_{\theta }$, from which the predictive mean $\boldsymbol{\mu}_\theta$ and quantile intervals for ${\mathbf{Y}}_{{{\mathcal{T}}_{2}},\mathcal{V}}$ are obtained. Under the assumption of Gaussian observation noise with variance $\sigma^{2}$, i.e., $y_{t,v}\mid\theta,\sigma^2 \sim \mathcal{N}\!\big(\mu_\theta(t,v),\,\sigma^2\big)$, the negative log-likelihood for a single observation is given by

\begin{equation}
\label{eq:nll}
\ell_{t,v}(\theta,\sigma^2)=\frac{1}{2\sigma^{2}}\big(y_{t,v}-\mu_\theta(t,v)\big)^2+\frac{1}{2}\log(2\pi\sigma^2)
\end{equation}

Aggregating over all spatio-temporal locations in a given prediction window, the loss function for model training can be written as:

\begin{equation}
\label{eq:loss2}
\mathcal{L}_{\text{NLL}}(\theta,\sigma^2)=\sum_{(t,v)\in\mathcal{O}}\ell_{t,v}(\theta,\sigma^2)
\end{equation}

\section{Methodology}
\label{methodology}
\subsection{Overall Framework}
\begin{figure*}
\centering
\includegraphics[width=0.9\textwidth]{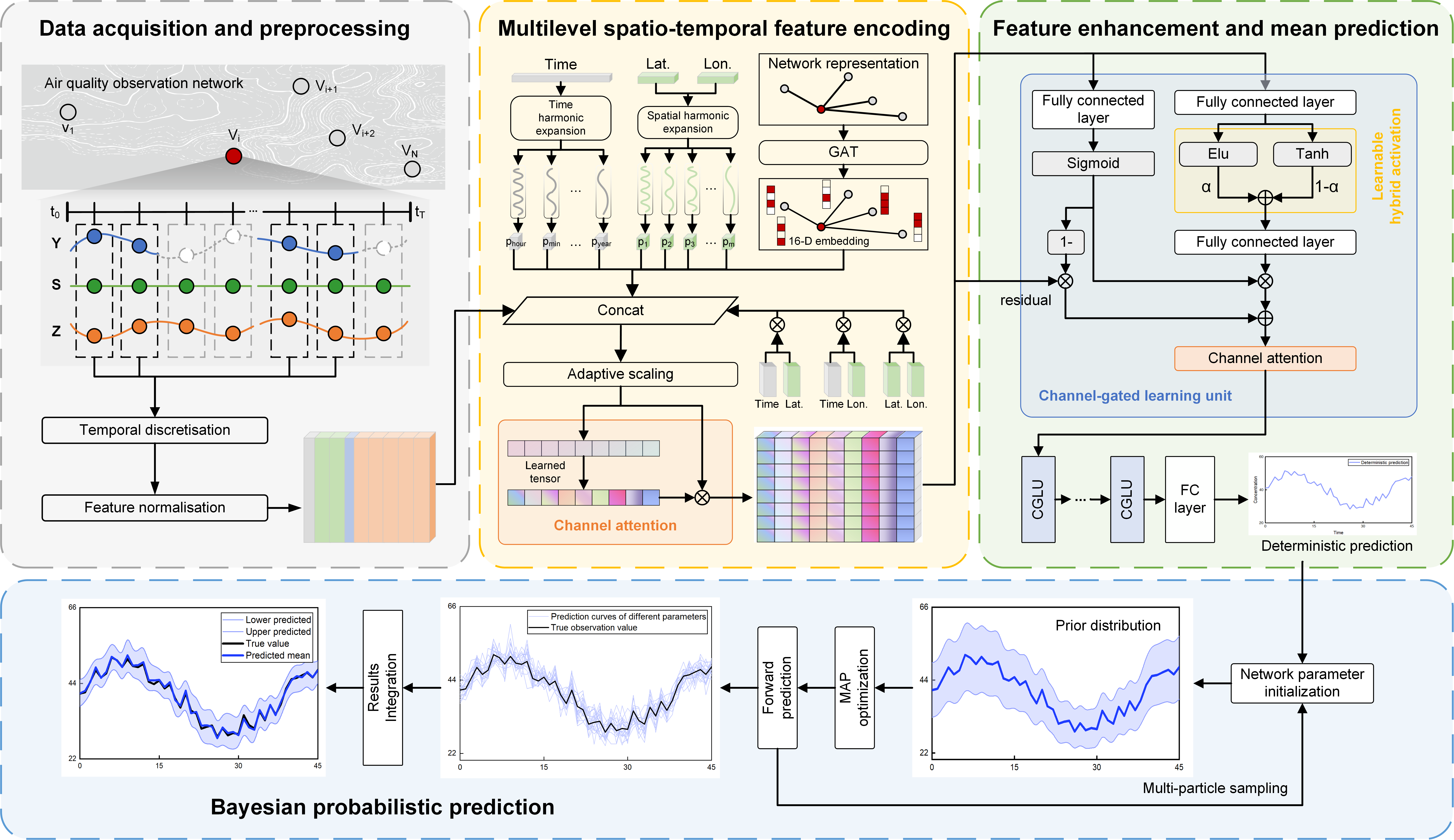}
\caption{The CGLU-BNF prediction framework architecture diagram.}
\label{fig4}
\end{figure*}

The CGLU-BNF framework captures and enhances long range spatio-temporal dynamics in incomplete observations, produces direct predictions under diverse missing data patterns, and supplies reliable uncertainty estimates. It comprises three key components: multilevel spatio-temporal feature encoding (MSFE), feature enhancement and mean prediction (FEMP), and Bayesian probabilistic prediction (BPP). Fig.\ref{fig4} presents the overall architecture of the proposed CGLU-BNF model for air quality prediction. 

First, the MSFE module builds a high dimensional spatio-temporal representation from historical observations and exogenous covariates. It combines temporal harmonics, spatial Fourier embeddings, and GAT to capture seasonal periodicities and cross site dependencies under irregular sampling. Second, the FEMP module employs multi-layer channel gated learning units to adaptively filter and enhance spatio-temporal features, mapping them to conditional means. Finally, the BPP module performs particle based maximum a posteriori (MAP) inference to estimate the predictive distribution and to produce point forecasts with confidence intervals.

CGLU-BNF takes incomplete historical sequences and auxiliary features as input and outputs per-node predictions and quantiles over the forecast horizon. This end-to-end design avoids pre-interpolation, accommodates multiple missingness patterns, and preserves computational efficiency and accuracy.

\subsection{Data Acquistion and Preprocessing}
To ensure stable training with incomplete observations, we adopt a three-step preprocessing pipeline: (i) sample filtering invalid records that lack target values. (ii) temporal discretization: timestamps are mapped to integer indices by calculating offsets from a reference time, with the minimum shifted to zero. (iii) feature normalization: apply z-score scaling to all non-temporal features to mitigate scale disparities. For each $(t,v)$ we assemble the model input by concatenating the normalized time features, the spatial coordinates, and exogenous variables.

After processing, each sample ${{\mathbf{f}}_{i}}$ is represented as:

\begin{equation}
\label{eq:Temporal discretisation}
{{{t}'}_{i}}=index({{t}_{i}})-index({{t}_{0}})
\end{equation}

\begin{equation}
\label{eq:Cleaned samples}
{{\mathbf{f}}_{i}}=[{{{t}'}_{i}},{{\mathbf{{s}'}}_{i}},{{{y}'}_{{{t}_{i}},{{v}_{i}}}},{{\mathbf{{Z}'}}_{{{t}_{i}},{{v}_{i}}}}]
\end{equation}

All processed samples are stacked to form the feature matrix ${{\mathbf{F}}_{prepro}}$:

\begin{equation}
\label{eq:Feature matrix after preprocessing}
{{\mathbf{F}}_{prepro}}=[\mathbf{f}_{1}^{T},\cdots ,\mathbf{f}_{\eta }^{T}]\in {{\mathbb{R}}^{\eta \times d_{prepro}}}
\end{equation}

Among them, $d_{prepro}$ is the total feature dimension after data preprocessing. ${{{t}'}_{i}}$ is the discretized time index, ${{\mathbf{{s}'}}_{i}}$ the normalized spatial coordinate vector, ${{{y}'}_{{{t}_{i}},{{v}_{i}}}}$ the normalized pollutant concentration, and ${{\mathbf{{Z}'}}_{{{t}_{i}},{{v}_{i}}}}$ the corresponding vector of normalized exogenous covariates.

\subsection{Multilevel Spatio-temporal Feature Encoding}
Conventional spatio-temporal encoders face three main limitations: (i) conflating seasonal periodicity with long term trends; (ii) relying on position invariant spatial kernels; and (iii) assigning equal importance to all features. These issues are exacerbated when observations are sparse or unevenly distributed. To overcome them, we build a multilevel spatio-temporal feature encoder that converts raw and incomplete inputs into a unified high dimensional representation while preserving the temporal seasonality and spatial correlations that govern air quality dynamics.

Specifically, we retain the preprocessed features $\mathbf{F}_{prepro}$ and augment them with spatio-temporal interaction terms ${\mathbf{F}}_{ts}$ and purely spatial longitude–latitude products ${\mathbf{F}}_{ss}$. Temporal harmonics are introduced to capture multiscale seasonality, and spatial harmonics to model smooth geographic gradients. The GAT generates site-level attention embeddings that provide neighborhood context. Each feature block is assigned a learnable scaling coefficient optimize together with the network weights. All features are concatenated, and a channel attention layer rescales their magnitudes before they enter the feature enhancement and mean prediction module. This integrated structure disentangles linear, periodic, and local dependencies, automatically balances their scales and saliencies, and yields numerically stable, informative representations robust to incomplete spatio-temporal sampling.

\subsubsection{Interaction Terms}
In spatio-temporal dynamic modeling, temporal or spatial characteristics alone often cannot fully characterize the nonlinear evolution of pollutants. Therefore, we explicitly introduce spatio-temporal interaction and the purely spatial interaction. Spatio-temporal interaction terms are used to characterize the dynamic dependencies inherent in the temporal evolution of the same location. Their mathematical form can be expressed as:

\begin{equation}
\label{eq:spatio-temporal interaction}
{{\mathbf{F}}_{\text{ts}}}=(\mathbf{T}\odot {{\mathbf{S}}_{latitude}})\oplus (\mathbf{T}\odot {{\mathbf{S}}_{longitude}})
\end{equation}

On the other hand, the space-space interaction term can implicitly capture nonlinear geographic correlations in low dimensional spatial coordinates, thereby enhancing the ability to characterize complex diffusion patterns and regional differences. Specifically, for spatial vectors, we consider the interaction between their two dimensions:

\begin{equation}
\label{eq:space-space interaction}
{{\mathbf{F}}_{\text{ss}}}={{\mathbf{S}}_{latitude}}\odot {{\mathbf{S}}_{longitude}}
\end{equation}

\subsubsection{Temporal Seasonality Terms (TST)}
To explicitly represent seasonality across minutes to years and decouple it from long term trends, we introduce harmonic time features. We use orthogonal sine–cosine pairs as fixed bases that integrate smoothly into gradient based training. This construction captures multiple periods without adding trainable parameters to the basis itself. Let $P=\{{{p}_{1}},\cdots ,{{p}_{L}}\}$ denote the set of base periods, and for each $p_{l}$ define the harmonic orders ${{H}_{{{p}_{l}}}}=\{1,\cdots ,H_{{{p}_{l}}}^{\max }\}$ (with $H_{{{p}_{l}}}^{\max }\le \left| {{p}_{l}}/2 \right|$). Given the reindexed time ${{{t}'}_{i}}$, the $h$-th harmonic for period $p_{l}$ is

\begin{equation}
\label{eq:time period}
{{\xi }_{l,h}}({{{t}'}_{i}})=[\cos 2\pi h{{{t}'}_{i}}/{{p}_{l}},\sin 2\pi h{{{t}'}_{i}}/{{p}_{l}}]
\end{equation}

Then, all harmonics are concatenated by channel to form the seasonal characteristic term ${{\mathbf{F}}_{seasonality}}$.

\subsubsection{Spatial Fourier Terms (SFT)}
To mitigate the low frequency bias of deep networks and resolve multi-scale spatial structure at the city block scale\cite{tancik2020fourier}, we apply Fourier feature mapping to the normalized coordinates $\mathbf{{s}'}$. For axis $c\in\{\text{longitude},\text{latitude}\}$, define the harmonic orders $K_{c}=\{0,1,\cdots ,{{K}_{c-1}}\}$ and map the scalar coordinate $s'_{c}$ to

\begin{equation}
\label{eq:Fourier map}
{{\psi }_{c}}({{{s}'}_{c}})={{[\cos (2\pi {{2}^{k}}{{{s}'}_{c}}),\sin (2\pi {{2}^{k}}{{{s}'}_{c}})]}_{k\in {{K}_{c}}}}
\end{equation}

With the full spatial embedding ${{\mathbf{F}}_{spatial}}$. This yields multi frequency sine–cosine channels that enhance the resolution of spatial patterns.

\subsubsection{Spatial Aggregation Terms}
SFT mitigate low frequency bias and recover stationary, translation invariant structure. However, they do not encode network topology or flow direction, limiting their ability to model neighborhood specific, nonstationary couplings common in urban air quality (e.g., local advection, street canyon effects).

To address this limitation, we augment SFT with a single layer, multi-head graph attention network. The GAT preserves a distance-decay prior while learning direction-aware edge weights from data, thereby adapting to local topology and nonstationary dependencies. It also aggregates information from adjacent and multi-hop neighbors, improving robustness to sparse or irregular sampling. In this hybrid design, SFT supplies a global, frequency rich basis for stable learning of large scale, quasi-stationary patterns, whereas GAT injects adaptive local structure and directionality to capture anisotropy and nonstationary couplings.

For node $i$, we form a static site descriptor $\mathbf{h}_{i}^{(0)}$ from robust statistics (the long term mean and the 25th/75th percentiles), which is resilient to missingness and outliers. $O$-head GAT propagates messages only along edges with nonzero entries in $\mathbf{A}$, and its attention weights are adjusted by a Gaussian connectivity prior. The attention is

\begin{equation}
\label{eq:GAT attention}
\small
\left\{
\begin{aligned}
	e_{ij}^{(o)} &= \text{LeakyReLU}\left({\mathbf{a}^{(o)\top}} \left[{\mathbf{W}^{(o)}}\mathbf{h}_{i}^{(0)} \parallel {\mathbf{W}^{(o)}}\mathbf{h}_{j}^{(0)}\right]\right) \\
	\bar{e}_{ij}^{(o)} &= A_{ij} \cdot e_{ij}^{(o)} \\
	a_{ij}^{(o)} &= \operatorname{softmax}_{j\in N(i)} \left(\bar{e}_{ij}^{(o)}\right) \\
	\mathbf{h}_{i}^{(o)} &= \text{ELU} \left(\sum_{j\in N(i)} a_{ij}^{(o)} {\mathbf{W}^{(o)}} \mathbf{h}_{j}^{(0)}\right)
\end{aligned}
\right.
\end{equation}

Outputs are concatenated and linearly projected:

\begin{equation}
{{\mathbf{g}}_{i}}={{\mathbf{W}}_{out}}\text{concat}(\mathbf{h}_{i}^{(1)},\cdots ,\mathbf{h}_{i}^{(O)})
\end{equation}

where $N(i)=\{j: A_{ij}>0\}$. The learnable projection matrix of the $O$-th head is ${{\mathbf{W}}^{(o)}}$, and ${{\mathbf{a}}^{(o)}}$ is the associated attention kernel. Aggregating $\mathbf{g}_{i}$ over time yields the spatial feature $\mathbf{F}_{GAT}$.

\subsubsection{Adaptive Scaling and Channel Reweighting}
Spatio-temporal encoding yields a batch feature matrix $\mathbf{{F}'}\in {{\mathbb{R}}^{B\times M}}$, where $B$ is the batch size and $M$ is the channel count. 

\begin{equation}
\label{eq:Feature concat}
\small
\mathbf{{F}'}={{\mathbf{F}}_{prepro}}\oplus {{\mathbf{F}}_{ts}}\oplus {{\mathbf{F}}_{ss}}\oplus {{\mathbf{F}}_{seasonality}}\oplus {{\mathbf{F}}_{spatial}}\oplus {{\mathbf{F}}_{GAT}}
\end{equation}

The concatenated spatio-temporal features vary widely in amplitude, variance, and correlation. High amplitude channels dominate the gradients, weak signals are obscured, and redundant or noisy channels hinder efficiency and generalization. To counter these effects, the encoder employs two weighting mechanisms: a learnable adaptive scaling layer and a channel attention gate. The scaling layer automatically adjusts each feature’s magnitude during training, whereas the attention gate models inter channel dependencies, amplifies salient information, and suppresses redundancy. And placing scaling before attention avoids gate saturation and lets attention focus on information rather than raw magnitude.

Let $\boldsymbol{\zeta}$ be a set of learnable scaling coefficients. We scale each channel by broadcasting $\exp (\boldsymbol{\zeta })\in {{\mathbb{R}}^{M}}$ in the batch, achieving automatic rescaling of feature scales.

\begin{equation}
\label{eq:Feature Scaling}
{\mathbf{F}_{scale}}={{e}^{{{\boldsymbol{\zeta }}^{\text{T}}}}}\odot \mathbf{{F}'}
\end{equation}

The channel attention gate first applies global average pooling to the scaled features $\mathbf{F}_{scale}$, extracting channel statistics and removing spatial bias. 

\begin{equation}
\label{eq:pooling}
\mathbf{z}=\frac{1}{B}\sum\limits_{b=1}^{B}{{{\mathbf{F}}_{scale}}[b,:]}
\end{equation}

A two-layer fully connected network then produces a gating vector $\mathbf{w}_{ca}$ that captures nonlinear inter channel dependencies. 

\begin{equation}
\label{eq:weight vector1}
{{\mathbf{w}}_{ca}}=Sigmoid(\text{ReLU}(\mathbf{W}_{ca}^{1,1}\mathbf{z}+\mathbf{b}_{ca}^{1,1})\mathbf{W}_{ca}^{1,2}+\mathbf{b}_{ca}^{1,2})
\end{equation}

The final re-weighted representation is $\mathbf{F}_{ca}$.

\begin{equation}
\label{eq:weight vector2}
\mathbf{F}_{ca}=\mathbf{F}_{scale} \odot {{\mathbf{w}}_{ca}}^{\top}
\end{equation}

\subsection{Feature Enhancement and Mean Prediction}
In settings with incomplete spatio-temporal observations, multilevel feature encoding yields unified representations but still contains noise and imbalance from missing data and channel heterogeneity. To suppress this noise, emphasize critical signals, and map high dimensional features to pollutant concentrations robustly, we add a feature enhancement and mean prediction layer. Let ${{\mathbf{h}}^{(l)}}$ denote the input to the $l$-th channel gated learning unit (CGLU). Each CGLU comprises a residual block with a learnable mixture activation and channel wise attention, followed by a lightweight MLP that produces the conditional mean.

To mitigate vanishing gradients, we employ a soft gated residual architecture that regulates information flow. The gates suppress noise while maintaining stable gradient propagation. Within each residual block, we first compute intermediate features via a linear transformation.

\begin{equation}
\label{eq:GRN1}
{{\mathbf{f}}^{(l)}}=\text{ELU}(\mathbf{W}_{1}^{(l)}{{\mathbf{h}}^{(l)}}+\mathbf{b}_{1}^{(l)})\mathbf{W}_{2}^{(l)}+\mathbf{b}_{2}^{(l)}
\end{equation}

Here, ${{\mathbf{h}}}^{(l)}$ denotes the input feature vector at layer $l$; $\mathbf{W}^{(l)}$ and $\mathbf{b}^{(l)}$ are the linear projection weights and bias vector of the gated residual subnetwork. The gating vector ${{\boldsymbol{\gamma}}^{(l)}}$ is computed to adaptively fuse the original and enhanced features, where ${\mathbf{W}_{g}^{(l)}}$ and ${\mathbf{b}_{g}^{(l)}}$ are the parameters used to generate ${{\boldsymbol{\gamma }}^{(l)}}$.

\begin{equation}
\label{eq:GRN2}
{{\boldsymbol{\gamma }}^{(l)}}=\text{Sigmoid}(\mathbf{W}_{g}^{(l)}{{\mathbf{h}}^{(l)}}+\mathbf{b}_{g}^{(l)})
\end{equation}

Note that conventional residual or deep networks fix the activation function (e.g., ReLU or ELU). Such rigid nonlinearities limit expressiveness and can saturate when the input distribution shifts, especially at high missing rates. To enable adaptive selection of nonlinearities and refine hidden representations during training, we replace the fixed activation function with a trainable convex combination of ELU and Tanh. This design enhances both model fitting and uncertainty quantification. ELU preserves negative responses and accelerates convergence, whereas Tanh provides smooth, bounded outputs. The resulting learnable activation is defined as follows:

\begin{equation}
\label{eq:GRN3}
{{\phi }_{{{\alpha }^{(l)}}}}(\mathbf{u})={{\alpha }^{(l)}}\text{ELU}(\mathbf{u})+(1-{{\alpha }^{(l)}})\text{Tanh}(\mathbf{u})
\end{equation}

The hybrid activation ${{\phi }_{{{\alpha }^{(l)}}}}$ learns a convex mixture of ELU and Tanh via the trainable coefficient ${{\alpha }^{(l)}} \in[0,1]$. Together with the gating mechanism, it yields the gated residual output $\mathbf{r}^{(l)}$:

\begin{equation}
\label{eq:GRN4}
{{\mathbf{r}}^{(l)}}=(1-{{\boldsymbol{\gamma }}^{(l)}})\odot {{\mathbf{h}}^{(l)}}+{{\boldsymbol{\gamma }}^{(l)}}\odot {{\phi }_{{{\alpha }^{(l)}}}}({{\mathbf{f}}^{(l)}})
\end{equation}

To enhance discriminative spatio-temporal feature selection, we incorporate channel attention to adaptively recalibrate each channel. As in Eqs.\ref{eq:pooling}–\ref{eq:weight vector2}, global average pooling followed by a two-layer MLP produces the attention weights $\boldsymbol{\omega }_{\text{ca}}^{(l)}$, which are then used to reweight the channels.

\begin{equation}
\label{eq:GRN5}
{{\mathbf{\hat{h}}}^{(l)}}={{\mathbf{r}}^{(l)}}\odot \boldsymbol{\omega }_{\text{ca}}^{(l)}
\end{equation}

After $L$ stacked CGLUs, the conditional mean is produced by a lightweight MLP.

\begin{equation}
\label{eq:GRN6}
{{\boldsymbol{\mu }}_{\theta }}={{\mathbf{W}}_{mlp}}{{\mathbf{\hat{h}}}^{(l)}}+{{\mathbf{b}}_{mlp}}
\end{equation}

where ${\boldsymbol{\mu }_{\theta }}$ denotes the point estimate, and $\mathbf{W}_{mlp}$ and $\mathbf{b}_{mlp}$ are the MLP weight matrix and bias vector.

\subsection{Bayesian Probabilistic Prediction}
We append a Bayesian output layer to the deterministic backbone to quantify predictive uncertainty under missing and noisy observations. We instantiate the conditional prediction model in Eq.\ref{eq:mapping function} using a Gaussian likelihood function and weakly informative logistic prior. Given the spatio-temporal input matrix ${\mathbf{X}}$, the FEMP module provides the conditional mean ${\boldsymbol{\mu }_{\theta }}(X)$. Assuming i.i.d. Gaussian observation noise with variance ${{\sigma }^{2}}$, the likelihood is

\begin{equation}
\label{eq:BPP1}
\mathbf{Y} \mid \boldsymbol{\theta}, \sigma^{2}
\sim 
\mathcal{N}\big(
\boldsymbol{\mu}_{\theta}(\mathbf{X}),
\sigma^{2}\mathbf{I}_{\eta}
\big)
\end{equation} 

Where, $\sigma$ denotes the standard deviation of the observation noise; $\boldsymbol{\theta}$ collects all network parameters (e.g. scaling coefficients, weights, biases, and activation mixing coefficients); and $\eta$ is the number of observed entries contributing to the training loss in Eqs.\ref{eq:nll}–\ref{eq:loss2}.

We place independent $\mathrm{Logistic}(0,1)$ priors on $\log \sigma $ and on each component of $\boldsymbol{\theta}$ to mitigate overfitting and enhance robustness. Parameters are estimated via multiple MAP optimizations initialized from different starting points.

\begin{equation}
\label{eq:BPP2}
(\boldsymbol{\theta}^{*}, \sigma^{*})
\in 
\arg\!\underset{\boldsymbol{\theta},\,\sigma}{\max}\,
\Big[
\log \rho(\mathbf{Y}\mid \mathbf{X}, \boldsymbol{\theta}, \sigma)
+ 
\log \pi(\boldsymbol{\theta}, \sigma)
\Big]
\end{equation} 

By running MAP optimization from multiple random initializations, we obtain $M$ local modes $\{({{\boldsymbol{\theta }}_{m}},{{\sigma }_{m}})\}_{m=1}^{M}$. To enhance stability and convergence, training uses Adam with a cosine-annealed learning rate and global gradient clipping. At test time, each solution $(\boldsymbol{\theta}_{m},\sigma_m)$ yields a Gaussian predictive component $\mathcal{N}(\boldsymbol{\mu} _{*}^{(m)},\sigma _{*}^{2(m)})$ for a new input. An equal weight mixture of these components approximates the posterior predictive distribution:

\begin{equation}
\label{eq:BPP3}
\rho(\mathbf{y}_{*}\mid \mathbf{x}_{*}, \mathcal{D})
\approx
\frac{1}{M}\sum_{m=1}^{M}
\mathcal{N}\!\big(
\mathbf{y}_{*}\mid
\boldsymbol{\mu}_{*}^{(m)},\, \sigma_{*}^{2(m)}\mathbf{I}
\big)
\end{equation}

Here, $\boldsymbol{\mu}_{*}^{(m)}$ and $\sigma _{*}^{2(m)}$ denote the predictive mean and the observation noise variance of the $m$-th posterior mode, respectively.

The final output mean is $\mathbb{E}[\mathbf{y}_{*}] = 
\frac{1}{M}\sum_{m=1}^{M}\boldsymbol{\mu}_{*}^{(m)}$, which corresponds to the prediction uncertainty:
\begin{equation}
\label{eq:BPP4}
\operatorname{Var}(\mathbf{y}_{*})
=
\frac{1}{M}\sum_{m=1}^{M}\sigma_{*}^{2(m)}
+\operatorname{Var}_{m}\!\big(\boldsymbol{\mu}_{*}^{(m)}\big)
\end{equation}

\section{Experiments}
\label{Experiments}
This section assesses the CGLU-BNF framework for air quality prediction under multiple data missing scenarios.

\subsection{Dataset and Configurations}
\subsubsection{Dataset Description}
\begin{figure}[htpb!]
	\centering
	\includegraphics[width=0.9\columnwidth]{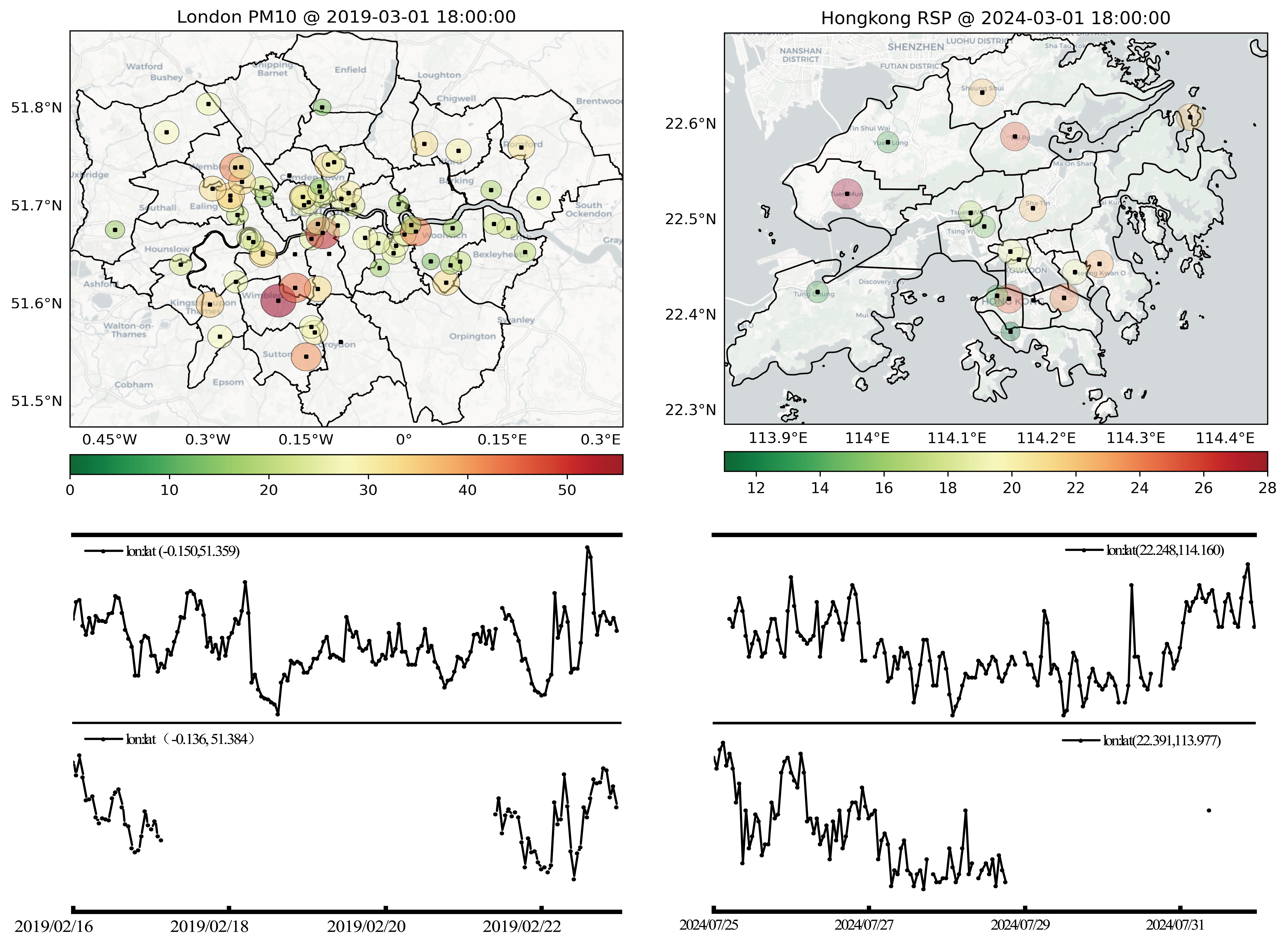}
	\caption{Slices of spatial and temporal observations of air quality datasets. The first row shows spatial slices of air quality across monitoring stations in London and Hong Kong at a fixed time. The second row presents temporal slices of complete air quality time series at a representative station in each city. The third row displays temporal slices of sparse air quality observations at the same stations and corresponding time periods.}
	\label{fig5}
\end{figure}

To quantitatively assess CGLU-BNF’s predictive performance under incomplete observations, we conduct $PM_{10}$ forecasting experiments on two publicly available, large scale air quality datasets. The two datasets are distributed in different regions with different missing rates and can cover a variety of complex scenarios. Table \ref{tab2:Data description} lists the detailed information of the datasets, and Fig.\ref{fig5} shows their spatio-temporal observation snapshots, which visualize the nonstationarity and periodicity of the air quality data and other statistical features. Notably, the London dataset includes only spatio-temporal attributes (time, latitude and longitude) and $PM_{10}$ concentrations, whereas the Hong Kong dataset additionally provides exogenous covariates($SO_{2}$, $NO_{2}$, $O_{3}$, $PM_{2.5}$). Furthermore, we do not impute missing ground truth values; instead, we exclude them during evaluation so that all models are assessed on the same set of observed targets.

\begin{table*}[!htbp]
	\centering
	\caption{Data description}
	\label{tab2:Data description}
	\renewcommand{\arraystretch}{1.3}
	\begin{tabularx}{\textwidth}{ccccccccl}
		\hline
		\textbf{Datasets} & \textbf{Regions} & \textbf{Frequency} & \textbf{Time span}         & \textbf{Nodes} & \textbf{Time points} & \textbf{Observations} & \textbf{Missing rate}  \\ \hline
		Air quality1\cite{hamelijnck2021spatio} & London          & Hourly          & 2018-12-31 to 2019-03-31 & 72            & 2161         & 155592         & 7.32\%     \\
		Air quality2\cite{HK_AirBase} & Hong Kong          & Hourly           & 2023-01-01 to 2024-12-31 & 18            & 17544        & 315792         & 3.09\%     \\
		\hline
	\end{tabularx}
\end{table*}

\subsubsection{Experimental Setting}
To assess robustness across models and varying degrees of incompleteness, we simulate four common missing data scenarios. (i) random missing, where values are lost sporadically in the data stream (e.g., packets dropped during transient communication degradation); (ii) node missing, where all observations from a single node are absent for an extended period (e.g., continuous sensor failure); (iii) timestamp missing, where data from every node are simultaneously unavailable at a specific time (e.g., a localized power outage); and (iv) block missing, where gaps form contiguous spatio-temporal blocks (e.g., a moving sensor passing through a tunnel that creates a persistent blind spot). For each scenario, the missing rate is varied from 10\% to 80\% in 10\% increments. And target values are removed according to the specified missing pattern. For cross validation, observation sites are randomly partitioned into five disjoint subsets. In each fold, the last month of records from sites in the held-out subset constitutes the test set. The training set includes all remaining data: full period observations from the other sites and non-test periods from the held-out sites.

All experiments were conducted on a server equipped with an Intel Xeon Gold 6133 CPU and four NVIDIA GeForce RTX4090 GPUs. The CGLU-BNF model comprises three stacked channel-gated learning layers, each with 512 hidden units and 16 particles. Training uses the AdamW optimizer with an initial learning rate of $5\times {{10}^{-3}}$, a batch size of 512, and a maximum of 5000 epochs. Hyperparameters were first tuned in preliminary trials, and the contribution of each module was assessed through ablation studies. Under identical settings, CGLU-BNF and all baseline models were trained and validated on five non-overlapping splits of each dataset, and average performance metrics were reported to enable a fair comparison under incomplete observation scenarios.

\subsubsection{Baselines}
To benchmark the CGLU-BNF framework under incomplete data conditions, we compare it against five baseline models on two public datasets. Baselines comprise classical statistical and machine learning predictors (HA, RF, STGBOOST) and end-to-end Gaussian process based methods that produce confidence intervals (ST-SVGP, BayesNF).

\begin{itemize}
	\item Historical Average (HA). This baseline computes the mean hourly pollutant concentration at each node from historical data and uses this constant value to forecast all future time steps.
	\item Random Forest (RF)\cite{breiman2001random}. An ensemble of decision trees is trained on bootstrap samples with randomly selected feature subsets, and their outputs are averaged, enabling robust spatio-temporal forecasting and effective modeling of nonlinear relationships.
	\item Spatio-Temporal Gradient Boosting Trees (STGBOOST)\cite{pedregosa2011scikit}. This extension of gradient boosting trees adapts the algorithm to spatio-temporal data, using recursive partitioning to capture nonlinear interactions between spatial and temporal factors and thus improve predictive accuracy.
	\item Spatio-Temporal Sparse Variational Gaussian Process (ST-SVGP)\cite{hamelijnck2021spatio}. Employs a sparse variational Gaussian process with inducing points to handle high dimensional spatio-temporal data, enabling scalable, non-parametric predictions with quantified uncertainty.
	\item Bayesian Neural Fields (BayesNF)\cite{saad2024scalable}. Models high dimensional spatio-temporal function fields with Bayesian neural networks and employs MAP inference to deliver both predictive means and their associated uncertainty estimates.
\end{itemize}

\subsubsection{Performance Metrics}
To evaluate predictive accuracy and uncertainty quality, we report root mean square error (RMSE), mean absolute error (MAE), coefficient of determination ($R^{2}$), and symmetric mean absolute percentage error (SMAPE) for point forecasts, along with average interval width (AIW) and relative interval width mean (RIWM) for predictive intervals. The metrics are defined as follows:

\begin{equation}
	\label{eq: evaluate metrics}
	\left\{
	\begin{aligned}
	\mathrm{AIW}  &= \frac{1}{\eta }\sum\limits_{i=1}^{\eta }{({{\mu }_{\theta }}(i,upper)-{{\mu }_{\theta }}(i,lower))},\\
	\mathrm{RIWM} &= \frac{1}{\eta }\sum\limits_{i=1}^{\eta }{\frac{{{\mu }_{\theta }}(i,upper)-{{\mu }_{\theta }}(i,lower)}{{{y}_{i}}}}.
	\end{aligned}
	\right.
\end{equation}

\subsection{Experimental Results}
\subsubsection{Prediction Accuracy}
\begin{figure*}[t!]
	\centering
	\includegraphics[width=0.95\textwidth]{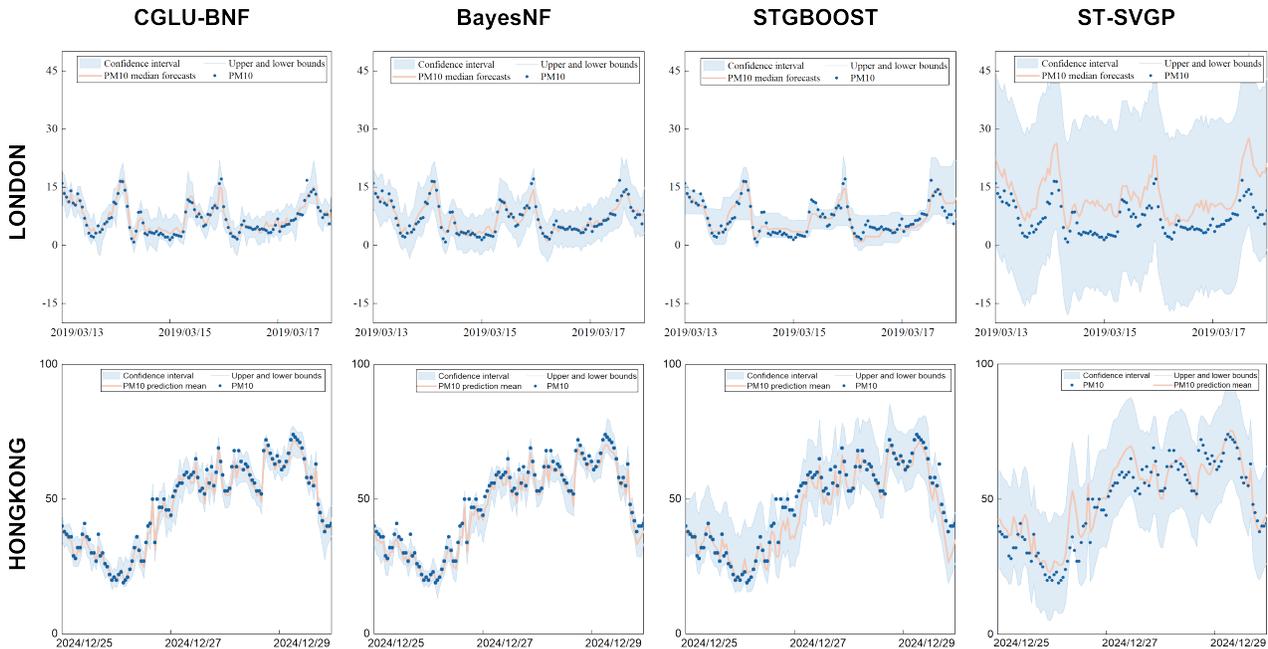}
	\caption{Predictive performance of the CGLU-BNF and baselines methods on the original London and Hong Kong datasets}
	\label{fig6}
\end{figure*}

Table \ref{tab3} summarizes predictive performance on both datasets, and Fig.\ref{fig6} plots predicted versus observed series for CGLU-BNF and the three uncertainty aware baselines. At low missing rates in original data, all methods benefit from dense observations; nevertheless, CGLU-BNF achieves the best point forecast accuracy and the sharpest prediction intervals among probabilistic models. On the London dataset, RMSE and MAE decrease to 7.26 $\mu g/{{m}^{3}}$ and 4.04 $\mu g/{{m}^{3}}$, yielding relative gains of 6.74\% and 7.95\% over the next best BayesNF. More importantly, its average interval width is only 11.86 $\mu g/{{m}^{3}}$, which is 18.85\% narrower than that of BayesNF. Consistent improvements are observed on the Hong Kong dataset, with improvements of 4.35\% and 5.63\% in RMSE and MAE, respectively.

\begin{table}[!htbp]
	\centering
	\caption{Results of the comparison of the prediction performance of the baseline model at the original missing rate for the two datasets. $R^2$ and SMAPE are reported in percentage (\%).}
	\label{tab3}
	\renewcommand{\arraystretch}{1.3}
	\resizebox{\columnwidth}{!}{
	\begin{tabular}{cccccccc}
		\hline
		Dataset                   & Method   & RMSE    & MAE     & $R^2$      & SMAPE   & AIW     & RIWM   \\ \hline
		\multirow{6}{*}{London}   & HA       & 15.67 & 11.01 & 0.09  & 48.27   & 0       & 0      \\
		& RF       & 10.37 & 6.09  & 0.61  & 28.64   & 0       & 0      \\
		& ST-SVGP  & 9.74  & 6.08  & 0.65  & 29.33   & 44.03 & 2.49 \\
		& STGBOOST & 8.80  & 5.12  & 0.72  & 24.35   & 25.45 & 1.15 \\
		& BayesNF  & 7.78  & 4.39  & 0.78  & 21.06   & 14.61 & 1.80 \\
		\rowcolor{gray!10}
		& CGLU-BNF & 7.26  & 4.04  & 0.81  & 19.47   & 11.86 & 1.64 \\ \hline
		\multirow{6}{*}{Hong Kong} & HA       & 25.11 & 20.19 & -1.21 & 52.37   & 0       & 0      \\
		& RF       & 4.81 & 3.63 & 0.92 & 8.61 & 0       & 0      \\
		& ST-SVGP  & 4.59 & 3.44 & 0.93 & 8.24 & 35.72 & 1.86 \\
		& STGBOOST & 4.38  & 3.22  & 0.93  & 7.74 & 16.59 & 0.81 \\
		& BayesNF  & 3.36  & 2.40  & 0.96  & 5.60  & 8.34 & 0.20 \\
		\rowcolor{gray!10}
		& CGLU-BNF & 3.22  & 2.27  & 0.96  & 5.24  & 8.07 & 0.19 \\ \hline
	\end{tabular}
}
\end{table}

Specifically, HA still yields large errors and an almost zero $R^{2}$, underscoring its inability to follow temporal fluctuations once the data depart from a smooth mean. RF reduces errors by a large amount, showing that nonlinear tree splits can exploit local covariates; however, its lack of explicit temporal and spatial modeling limits accuracy when dynamic dependencies and neighborhood correlations are present. ST-GBOOST narrows the gap by embedding coarse time lags into gradient boosted trees and produces plausible confidence intervals, yet hand-crafted lags cannot fully capture multiscale dependencies. ST-SVGP reproduces the overall trend but suffers from sizable point- and interval-prediction errors, likely because the Matérn kernel with separable spatio-temporal structure struggles with complex seasonality and nonstationarity, while manual selection of inducing points adds approximation bias. BayesNF improves performance by jointly learning Fourier-based temporal trends and spatial kernels, and by generating prediction intervals through a particle-based MAP head. However, because it lacks explicit spatial structure and feature selection mechanisms, local heterogeneity and noise are not sufficiently addressed. As a result, the model requires wider intervals to maintain adequate coverage.

CGLU-BNF delivers additional performance gains for two principal reasons. First, because the few remaining gaps are sparsely distributed when most sensors report normally, combined with the spatio-temporal feature coding layer of the graph attention can explicitly model nodes adjacency and data statistics. This design attenuates residual spatial patterns, tightens posterior spatial variance, and produces a smoother, more accurate reconstruction. Second, the channel gated learning unit adaptively re-weights feature maps, suppressing noisy channels while amplifying informative ones; its gated residuals connections preserve gradient flow, allowing a deeper network without overfitting and thus reducing both residual and model uncertainty.

\subsubsection{Random Missing Patterns}
\begin{figure*}[t!]
	\centering
	\includegraphics[width=0.95\textwidth]{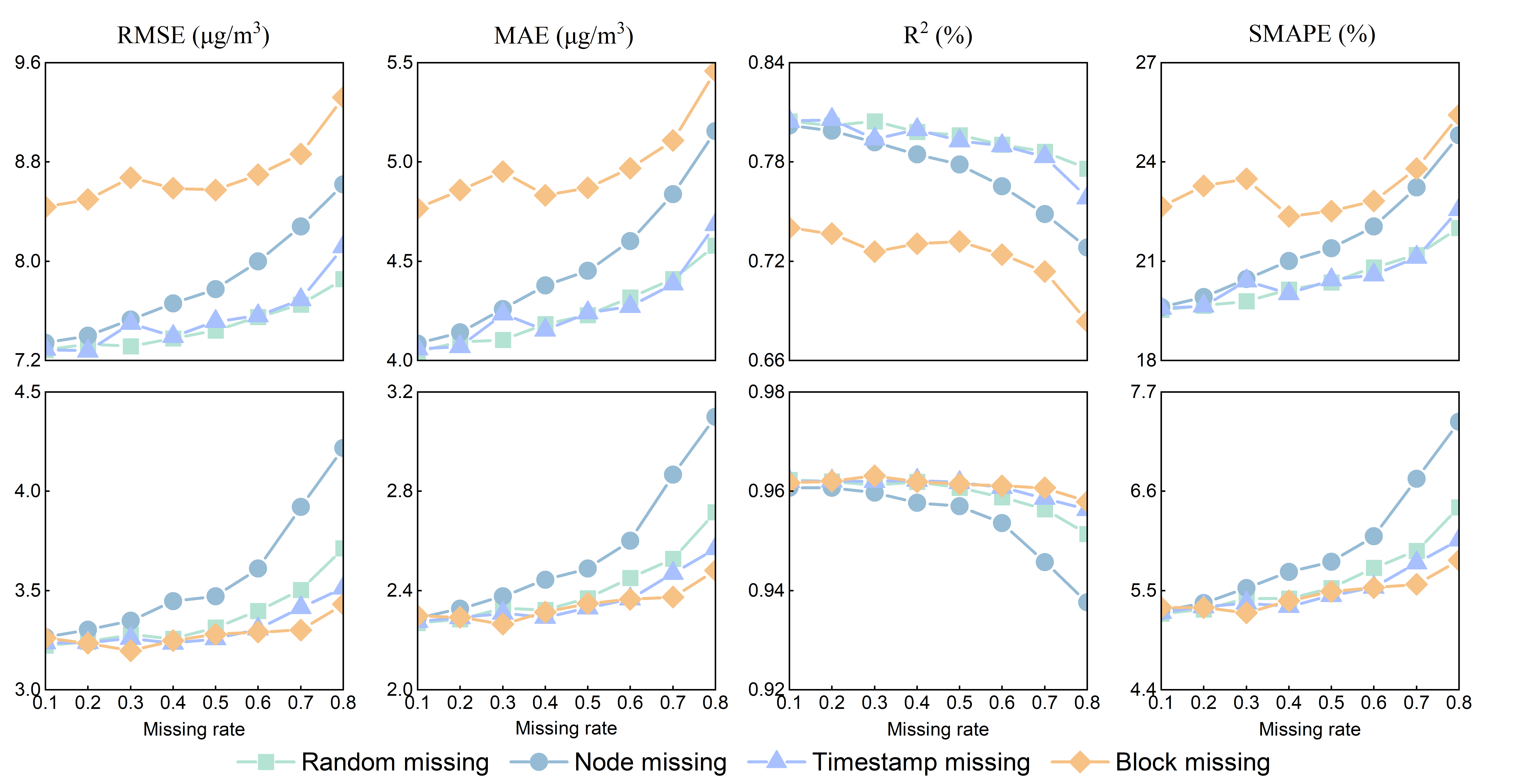}
	\caption{Predictive performance of CGLU-BNF under varying missing rates and missing data patterns. The first row shows the results for the London dataset, and the second row shows the results for the Hong Kong dataset.}
	\label{fig7}
\end{figure*}

Random missing in operational atmospheric networks typically stem from transient packet loss or brief sensor interference. These point-like voids, lacking fixed structure, offer a stringent test of model robustness and generalization. We evaluated CGLU-BNF under random missing rates from 10\% to 80\% (square-marked curves in Fig.\ref{fig7}). As missingness increases, performance declines slightly without a critical breakpoint. Specifically, on the London dataset, RMSE rises from 7.29 $\mu g/{{m}^{3}}$ to 7.86 $\mu g/{{m}^{3}}$. The Hong Kong dataset exhibits the same trend.

Under a representative 30\% random missing setting, CGLU-BNF achieved the best performance on all metrics and delivered the narrowest prediction intervals (Table\ref{tab4}). Across both datasets, it reduced RMSE and MAE by 4.49\% and 5.48\%, respectively, relative to the next-best model, BayesNF. Additionally, compared to the predictions based on the original data, the results offer two new insights. First, raising the random missing rate to 30\% inflates errors and interval widths for every model, underscoring the influence of missingness patterns; the deterioration is most pronounced for probabilistic methods such as ST-SVGP and BayesNF, whereas CGLU-BNF retains strong robustness. Second, the performance gap widens: CGLU-BNF’s RMSE is 25.10\% lower than ST-SVGP’s and 5.43\% lower than BayesNF’s. These gains indicate that CGLU-BNF’s dynamic channel re-weighting distinguishes genuine fluctuations from information gaps and prunes redundant uncertainty, while baseline models compensate for missing data by broadening their intervals.

\begin{table}[!htbp]
	\centering
	\caption{Predictive performance of different models in scenarios with 30\% random missing data}
	\label{tab4}
	\renewcommand{\arraystretch}{1.3}
	\resizebox{\columnwidth}{!}{
\begin{tabular}{cccccccc}
	\hline
	Dataset          & Method & RMSE & MAE & $R^2$ & SMAPE & AIW & RIWM \\ \hline
	\multirow{6}{*}{London}   & HA              & 15.71       & 11.05     & 0.08      & 48.41          & 0            & 0             \\
	& RF              & 10.38       & 6.11       & 0.61      & 28.65          & 0            & 0             \\
	& STGBOOST        & 8.81        & 5.13       & 0.72      & 24.39          & 27.65     & 1.43        \\
	& ST-SVGP         & 9.85        & 6.16       & 0.64      & 29.72          & 44.56      & 2.50        \\
	& BayesNF         & 7.80       & 4.44        & 0.78      & 21.31          & 14.76      & 1.70        \\
	\rowcolor{gray!10}
	& CGLU-BNF        & 7.38        & 4.14       & 0.80      & 19.91          & 11.83      & 1.27        \\ \hline
	\multirow{6}{*}{Hong Kong} & HA              & 25.11       & 20.20      & -1.21     & 52.39          & 0            & 0             \\
	& RF              & 4.85        & 3.66      & 0.92      & 8.69           & 0            & 0             \\
	& STGBOOST        & 4.43        & 3.26       & 0.93     & 7.84           & 16.74      & 0.82        \\
	& ST-SVGP         & 4.71             & 3.50            & 0.92          & 8.43              & 36.74      & 1.85        \\
	& BayesNF         & 3.40        & 2.43       & 0.96      & 5.67           & 8.29       & 0.20        \\
	\rowcolor{gray!10}
	& CGLU-BNF        & 3.28        & 2.33       & 0.96      & 5.41           & 7.54      & 0.19        \\ \hline
\end{tabular}
	}
\end{table}

\subsubsection{Node Missing}
Node missing occurs when an entire monitoring station is offline for an extended period (e.g., hardware failure or power outage), causing the simultaneous loss of all observations. We operationalize this as a 24-hour outage at a site on a given day. As shown by the circle-marked curve in Fig.\ref{fig7}, CGLU-BNF’s performance declines gradually as the missing rate increases from 10\% to 80\%. On the London dataset, RMSE and MAE rise by 1.28 $\mu g/{{m}^{3}}$ and 1.07 $\mu g/{{m}^{3}}$, respectively, indicating strong resilience to sporadic station failures. When outages become widespread, the graph structure sparsifies and long range paths shrink, substantially increasing the difficulty of spatial extrapolation.

\begin{table}[!htbp]
	\centering
	\caption{Predictive performance of different models in scenarios with 30\% node missing data}
	\label{tab5}
	\renewcommand{\arraystretch}{1.3}
	\resizebox{\columnwidth}{!}{
		\begin{tabular}{cccccccc}
			\hline
			Dataset          & Method & RMSE & MAE & $R^2$ & SMAPE & AIW & RIWM \\ \hline
			\multirow{6}{*}{London}   & HA              & 15.69         & 10.95        & 0.09        & 48.03          & 0            & 0             \\
			& RF              & 10.33         & 6.14         & 0.61        & 28.75          & 0            & 0             \\
			& STGBOOST        & 8.89          & 5.17         & 0.71        & 24.63          & 27.59        & 1.43          \\
			& ST-SVGP         & 9.85          & 6.18         & 0.64        & 29.90          & 44.08        & 2.47          \\
			& BayesNF         & 7.91          & 4.52         & 0.77        & 21.67          & 14.51        & 1.71          \\
			\rowcolor{gray!10}
			& CGLU-BNF        & 7.53          & 4.26         & 0.79        & 20.47          & 12.84        & 1.45          \\ \hline
			\multirow{6}{*}{Hong Kong} & HA              & 25.39         & 20.46        & -1.23       & 53.29          & 0            & 0             \\
			& RF              & 4.98          & 3.75         & 0.91        & 8.91           & 0            & 0             \\
			& STGBOOST        & 4.47          & 3.30         & 0.93        & 7.95           & 16.80        & 0.82          \\
			& ST-SVGP         & 4.67             & 3.55            & 0.92           & 8.64              & 36.72            & 1.87             \\
			& BayesNF         & 3.56          & 2.56         & 0.96        & 6.01           & 8.38         & 0.21          \\
			\rowcolor{gray!10}
			& CGLU-BNF        & 3.35          & 2.38         & 0.96        & 5.53           & 7.62         & 0.19          \\ \hline
		\end{tabular}
	}
\end{table}

Under the 30\% node missing scenario, CGLU-BNF remains the top performer (Table \ref{tab5}). On the London dataset, its RMSE and MAE are 7.53 $\mu g/{{m}^{3}}$ and 4.26 $\mu g/{{m}^{3}}$, improving over BayesNF by 5.35\% and 6.39\%, respectively; the Hong Kong dataset shows comparable gains with 3.35 $\mu g/{{m}^{3}}$ and 2.38 $\mu g/{{m}^{3}}$. Although probabilistic baselines capture spatial correlations, their lack of explicit spatial interpolation and limited local feature modeling cause marked degradation. By contrast, CGLU-BNF leverages multi-source spatial structure: a graph attention layer aggregates multi-hop information from neighboring stations when nodes are missing, while spatial Fourier embeddings provide global periodic bases that bridge local gaps and support long range extrapolation. Notably, most models perform worse under node missing than under random or timestamp missing, underscoring the need for fine grain inter-site dependency modeling to sustain accuracy.

\subsubsection{Timestamp Missing}
To evaluate temporal generalization under sudden data outages, we construct a timestamp missing scenario in which all stations simultaneously lack observations at specific moments. As shown by the triangle-marked curve in Fig.\ref{fig7}, CGLU-BNF degrades smoothly as the missing rate increases from 10\% to 80\%, with negligible variation at low rates. On the London dataset, RMSE and MAE rise by 0.83 $\mu g/{{m}^{3}}$ and 0.63 $\mu g/{{m}^{3}}$, respectively; on the Hong Kong dataset, they increase by 0.28 $\mu g/{{m}^{3}}$ and 0.29 $\mu g/{{m}^{3}}$. These results indicate that the model effectively leverages cross day periodicity to infer short- to medium-term gaps. Moreover, errors grow less than under node-level missingness because spatial information remains intact, allowing the model to exploit inter station correlations and maintain superior overall performance.

\begin{table}[!htbp]
	\centering
	\caption{Predictive performance of different models in scenarios with 30\% timestamp missing data}
	\label{tab6}
	\renewcommand{\arraystretch}{1.3}
	\resizebox{\columnwidth}{!}{
		\begin{tabular}{cccccccc}
			\hline
			Dataset          & Method & RMSE & MAE & $R^2$ & SMAPE & AIW & \textbf{RIWM} \\ \hline
			\multirow{6}{*}{London}   & HA              & 15.77         & 11.15        & 0.08        & 48.78          & 0            & 0             \\
			& RF              & 10.45         & 6.18         & 0.60        & 28.92          & 0            & 0             \\
			& STGBOOST        & 8.85          & 5.13         & 0.71        & 24.34          & 27.55        & 1.40          \\
			& ST-SVGP         & 9.79          & 6.09         & 0.65        & 29.40          & 47.26        & 2.45          \\
			& BayesNF         & 8.04          & 4.59         & 0.76        & 21.95          & 14.45        & 1.74          \\
			\rowcolor{gray!10}
			& CGLU-BNF        & 7.50          & 4.23         & 0.79        & 20.41          & 12.69        & 1.43          \\ \hline
			\multirow{6}{*}{Hong Kong} & HA              & 25.57         & 20.68        & -1.29       & 54.16          & 0            & 0             \\
			& RF              & 4.82             & 3.65           & 0.92           & 8.63              & 0            & 0             \\
			& STGBOOST        & 4.36             & 3.20            & 0.93           & 7.67              & 16.65            & 0.83             \\
			& ST-SVGP         & 4.57             & 3.55            & 0.93           & 8.31              & 41.08            & 1.98            \\
			& BayesNF         & 3.34          & 2.40         & 0.96        & 5.60           & 7.92         & 0.19          \\
			\rowcolor{gray!10}
			& CGLU-BNF        & 3.26          & 2.31         & 0.96        & 5.36           & 7.13         & 0.17          \\ \hline
		\end{tabular}
	}
\end{table}

Under a representative 30\% timestamp missing setting, CGLU-BNF outperforms all baselines on every metric (Table \ref{tab6}). On the London dataset, it reduces RMSE and MAE by 6.75\% and 7.80\% relative to BayesNF, and by 15.25\% and 17.54\% relative to STGBOOST. On the Hong Kong dataset, RMSE and MAE drop by 2.31\% and 3.70\% compared with BayesNF. This divergence arises because timestamp missing creates simultaneous, moment wide gaps across sites, demanding strong cross-moment information propagation and robust extraction of cyclical trends. Traditional statistical models lack explicit mechanisms for inter-temporal transfer, yielding low accuracy. Probabilistic baselines capture spatial structure but struggle to represent fine grain, long period temporal patterns, making whole moment gaps difficult to bridge. By contrast, CGLU-BNF’s Fourier time series decomposition explicitly models long range trends and multiscale seasonality, while the channel gated residual unit amplifies persistent periodic signals and suppresses short term noise. Together, these components enable superior performance even under complete temporal outages.

\subsubsection{Block Missing}
To evaluate robustness under simultaneous temporal and spatial outages, we simulate a spatio-temporal block missing scenario in which every station lacks a continuous 24-hour segment on a given day. The diamond-marked curve in Fig.\ref{fig7} shows that errors increase monotonically as the missing rate rises from 10\% to 80\%. On the London dataset, CGLU-BNF’s RMSE and MAE increase by 0.89 $\mu g/{{m}^{3}}$ and 0.69 $\mu g/{{m}^{3}}$, respectively; on the Hong Kong dataset, they rise by 0.17 $\mu g/{{m}^{3}}$ and 0.18 $\mu g/{{m}^{3}}$. Compared with the random, node, and timestamp missing settings, London exhibits the largest degradation under block missing, whereas Hong Kong shows the smallest. The likely cause is that block missing in London breaks both spatial connectivity and temporal continuity, forcing complex spatio-temporal extrapolation. By contrast, the Hong Kong dataset includes exogenous covariates (e.g., multiple pollutant concentrations) that remain observed, providing continuous conditioning and cross-pollutant constraints. These inputs reduce extrapolation difficulty and yield the lowest errors in this scenario.

\begin{table}[!htbp]
	\centering
	\caption{Predictive performance of different models in scenarios with 30\% block missing data}
	\label{tab7}
	\renewcommand{\arraystretch}{1.3}
	\resizebox{\columnwidth}{!}{
		\begin{tabular}{cccccccc}
			\hline
			Dataset  & Method & RMSE & MAE & $R^2$ & SMAPE & AIW & RIWM \\ \hline
			\multirow{6}{*}{London}   & HA              & 15.80         & 11.28        & 0.07        & 49.13          & 0            & 0             \\
			& RF              & 12.38         & 7.32         & 0.44        & 32.13          & 0            & 0             \\
			& STGBOOST        & 10.14         & 6.00         & 0.63        & 27.19          & 27.39        & 1.39          \\
			& ST-SVGP         & 11.38         & 7.27         & 0.53        & 34.06          & 47.49        & 2.38          \\
			& BayesNF         & 9.20          & 5.31         & 0.69        & 24.81          & 14.91        & 1.76          \\
			\rowcolor{gray!10}
			& CGLU-BNF        & 8.67          & 4.95         & 0.73        & 23.49          & 13.69        & 1.86          \\ \hline
			\multirow{6}{*}{Hong Kong} & HA              & 25.77         & 20.90        & -1.33       & 54.98          & 0            & 0             \\
			& RF              & 4.93          & 3.69         & 0.92        & 8.71           & 0            & 0             \\
			& STGBOOST        & 4.43          & 3.26         & 0.93        & 7.80           & 16.41        & 0.81          \\
			& ST-SVGP         & 4.74             & 3.49            & 0.92           & 8.08              & 40.95            & 1.99             \\
			& BayesNF         & 3.38          & 2.43         & 0.96        & 5.72           & 7.97         & 0.20          \\
			\rowcolor{gray!10}
			& CGLU-BNF        & 3.20          & 2.26         & 0.96        & 5.25           & 7.29         & 0.18          \\ \hline
		\end{tabular}
	}
\end{table}

At a representative 30\% spatio-temporal block missing rate (Table \ref{tab7}), CGLU-BNF remains superior to all baselines. On the London dataset, it achieves an RMSE of 8.67 $\mu g/{{m}^{3}}$ and an MAE of 4.95 $\mu g/{{m}^{3}}$, improving over BayesNF by 5.76\% and 6.78\% and over STGBOOST by 14.50\% and 17.50\%. The Hong Kong dataset shows the same pattern, with RMSE and MAE gains of 5.33\% and 7.00\% relative to BayesNF, and 27.77\% and 30.67\% relative to STGBOOST. These results indicate higher point forecast accuracy and sharper interval estimates even under block missing conditions.

\section{Discussion}
\label{discussion}
This section analyzes the impact of prediction task requirements and model structure on accuracy and robustness from four complementary perspectives, including the effect of prediction duration, ablation studies of structural modules, and the contribution of exogenous covariates.

\subsection{Impact Analysis of Predicted Duration}
To quantify how forecast horizon length affects accuracy and uncertainty, we evaluated the performance of CGLU-BNF under two observation conditions: the original London dataset and its counterpart with 30\% random missing values. Test sets were configured with time spans of 1 day, 7 days, 14 days, and 21 days.

As shown in Table \ref{tab8}, both observation settings exhibit an error curve that first increases and then decreases with the prediction time span, peaking on the seventh day. This non-monotonic pattern aligns with weekly cycles: phase mismatches amplify errors toward the end of the first week. Upon entering the second week, the seasonal components of the cycle become more readily captured and mutually offset by the model, leading to a contraction in short-term high-frequency errors. The average interval width exhibits the same trend.

\begin{table}[!htbp]
	\centering
	\caption{Predictive performance of different models facing different prediction time horizon in the original data and 30\% random missing data}
	\label{tab8}
	\renewcommand{\arraystretch}{1.3}
	\resizebox{\columnwidth}{!}{
		\begin{tabular}{cccccccccc}
			\hline
			& Data  & \multicolumn{4}{c}{Original}          & \multicolumn{4}{c}{30\% random missing} \\ \hline
			Model                    &       & 1 day   & 7 days  & 14 days & 21 days & 1 day    & 7 days  & 14 days & 21 days \\ \hline
			\multirow{7}{*}{Our}     & RMSE  & 7.22  & 7.85  & 7.61  & 7.30  & 7.49   & 7.98  & 7.64  & 7.40  \\
			& MAE   & 4.39  & 4.81  & 4.40  & 4.02  & 4.57   & 4.95  & 4.46  & 4.14  \\
			& $R^2$ & 0.87  & 0.82  & 0.79  & 0.80  & 0.87   & 0.81  & 0.79  & 0.79  \\
			& SMAPE & 14.30 & 14.62 & 16.63 & 20.08 & 14.80  & 15.03 & 16.91 & 20.41 \\
			& AIW   & 12.99 & 14.75 & 13.64 & 12.42 & 12.70  & 14.65 & 13.77 & 12.43 \\
			& RIWM  & 0.47  & 0.50  & 0.59  & 1.27  & 0.47   & 0.50  & 0.60  & 1.39  \\ \hline
			\multirow{7}{*}{BayesNF} & RMSE  & 8.32  & 8.43  & 7.88  & 7.81  & 8.36   & 8.49  & 7.93  & 7.82  \\
			& MAE   & 5.15  & 5.35  & 4.70  & 4.38  & 5.06   & 5.39  & 4.74  & 4.41  \\
			& $R^2$ & 0.83  & 0.80  & 0.78  & 0.77  & 0.83   & 0.79  & 0.78  & 0.77  \\
			& SMAPE & 16.58 & 16.35 & 17.92 & 21.70 & 16.10  & 16.47 & 18.07 & 21.71 \\
			& AIW   & 16.46 & 17.34 & 15.78 & 14.51 & 15.97  & 17.45 & 15.62 & 14.54 \\
			& RIWM  & 0.62  & 0.60  & 0.70  & 1.73  & 0.60   & 0.60  & 0.69  & 1.75  \\ \hline
		\end{tabular}
	}
\end{table}

Introducing 30\% random missingness increases errors at short and medium horizons; however, as the horizon lengthens, the error gap between settings narrows. This suggests that low frequency trends and multiple scale seasonality dominate long range forecasts, while the disruptive effect of random gaps diminishes. Across both observation settings and all horizons, CGLU-BNF outperforms the baselines, effectively capturing weekly cycles and remaining robust to random omissions. Specifically, on the original data the MAE improvement is 11.04\%, and under 30\% random missingness it is 8.17\%.

\subsection{Ablation Experiments}
\begin{figure}[!htbp]
	\centering
	\includegraphics[width=0.9\columnwidth]{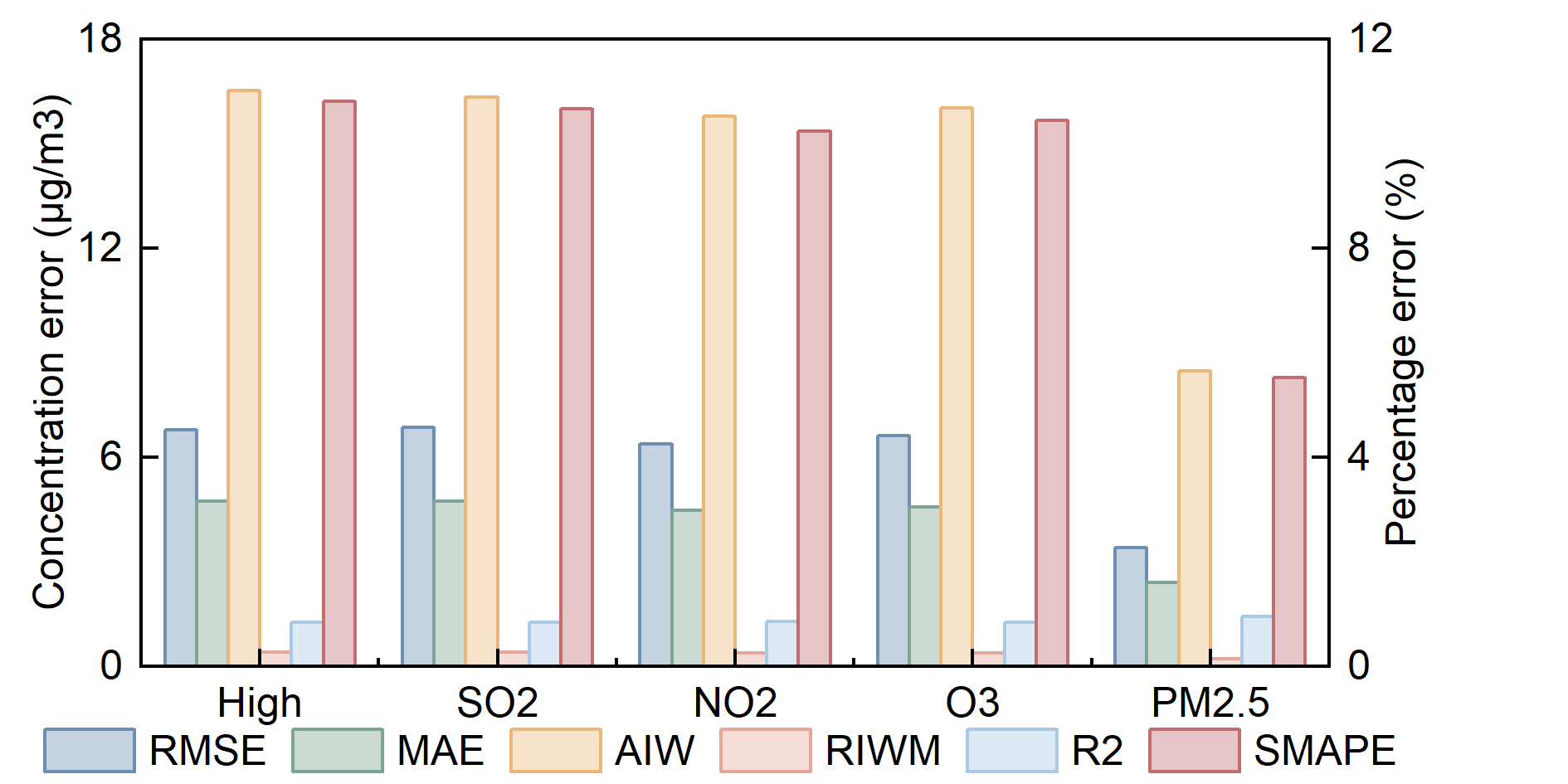}
	\caption{Performance comparison of ablation models on the London dataset.}
	\label{fig10}
\end{figure}

To assess the contribution of each model component, we designed ablation experiments using the London air quality dataset. To ensure comparability, all five model variants retained the training settings from the preceding section. Results are presented in Table \ref{fig10}.

\begin{itemize}
	\item No-TST: Temporal seasonal terms were removed from the Spatio-Temporal Feature Coding module;
	\item No-SFT: Spatial Fourier terms were removed from the Spatio-Temporal Feature Coding module;
	\item No-GAT: GAT is removed from the Spatio-Temporal Feature Coding module;
	\item No-CA: CA in the Concentration Inference module is removed;
	\item No-GRN: The GRN structure is replaced by the MLP network architecture;
	\item CGLU-BNF: The model is structurally complete.
\end{itemize}

The ablation results in Table \ref{fig10} reveal the relative contributions of each submodule within the model. Removing SFT increases RMSE by 10.6\% and doubles AIW, indicating that without the global spatial basis, the model struggles to reconstruct urban-scale long-wave gradients and high amplitude fluctuations, compensating passively by widening intervals through local smoothing. Removing TST increases error by 8.4\%, demonstrating the critical role of multiple scale periodic bases in capturing long term trends. Removing GRN caused MAE to rise by 7.3\% and AIW by 23\%, reflecting that insufficient deep nonlinear integration simultaneously amplifies mean bias and variance mismatch. Removing CA increases RMSE by 0.8\% and AIW by 0.18 $\mu g/{{m}^{3}}$, indicating that dynamic channel reweighting suppresses redundant features and enhances interval sharpness. Removing GAT increases RMSE by 0.7\%, indicating that multi-hop adjacency aggregation helps inject neighboring station anomalies and background gradients into target stations, thereby enhancing spatial extrapolation and robustness under structured missing data scenarios. Overall, CGLU-BNF achieves the smallest error and sharpest intervals while maintaining good coverage when retaining all components, demonstrating complementary synergy among submodules in balancing prediction accuracy and interval sharpness.

\subsection{Hyperparameter Sensitivity Analysis}
\begin{figure}[!htbp]
	\centering
	\includegraphics[width=0.9\columnwidth]{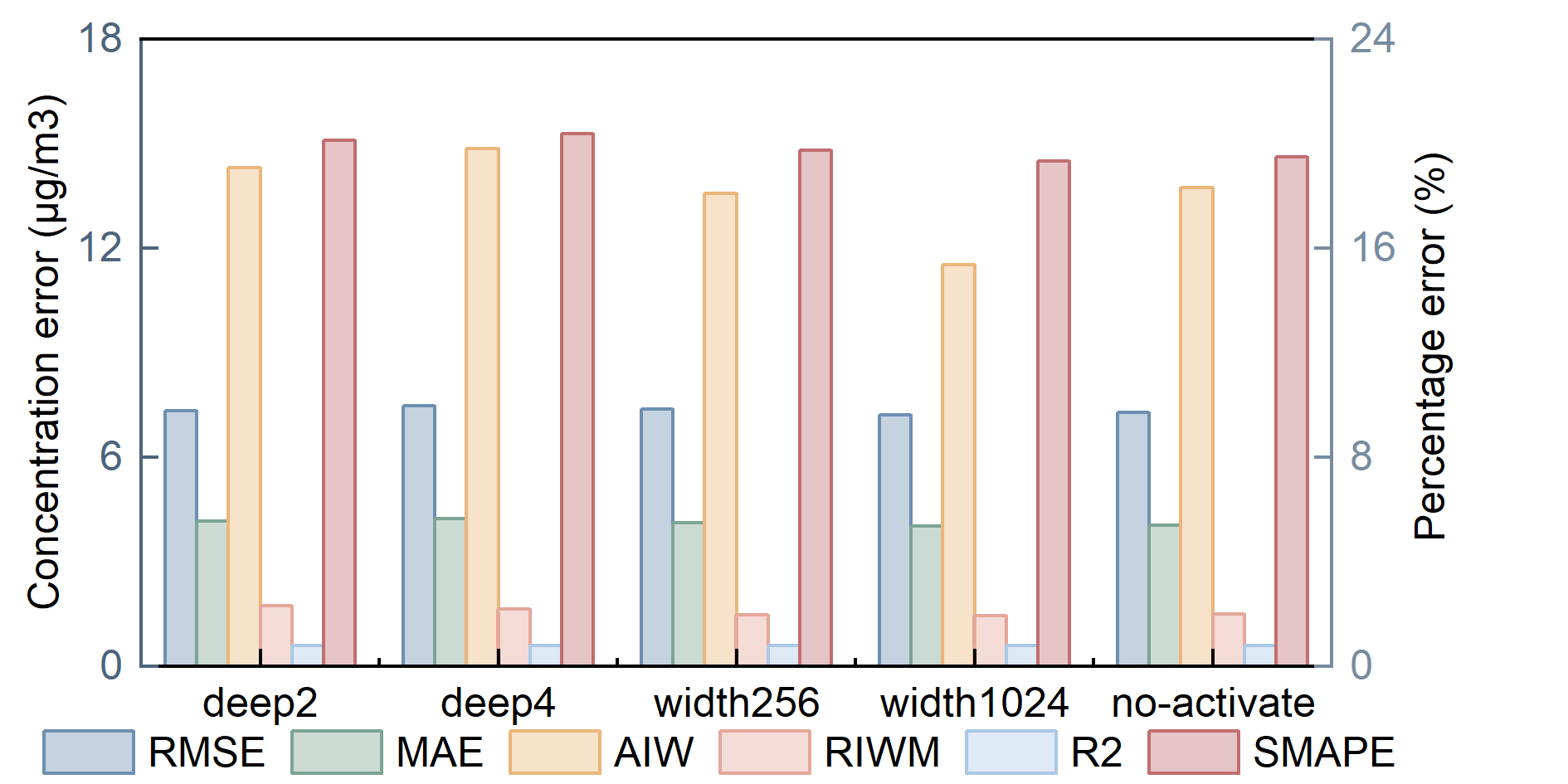}
	\caption{Effect of different hyperparameter settings on model prediction effectiveness.}
	\label{fig9}
\end{figure}

We also assessed the model’s hyperparameter sensitivity (Table \ref{fig9}) by varying network depth, width, and activation functions on the London dataset. Reducing the hidden layers from three to two (Deep-2) or increasing them to four (Deep-4) show a significant decrease in RMSE, MAE, $R^{2}$, and SMAPE, indicating that the three layer configuration already captures the essential spatio-temporal structure and that performance is relatively insensitive to depth. Expanding the hidden dimension from 256 to 1024 yields slight accuracy gains but incurs substantial training time and memory overhead, implying diminishing returns for wider networks when modelling high dimensional, sparse spatio-temporal dependencies. Replacing the learnable composite activation with a fixed, single activation modestly degrades all metrics, reaffirming that adaptive nonlinearities are valuable for extracting latent signals and improving forecast accuracy. Overall, a three layer architecture with 512 hidden units and a learnable activation mechanism offers a balanced trade-off between accuracy and computational efficiency while maintaining robust recovery of sparse spatio-temporal data.

\subsection{Analysis of $PM_{10}$ Concentration Impacts}
\begin{figure}[!htbp]
	\centering
	\includegraphics[width=0.9\columnwidth]{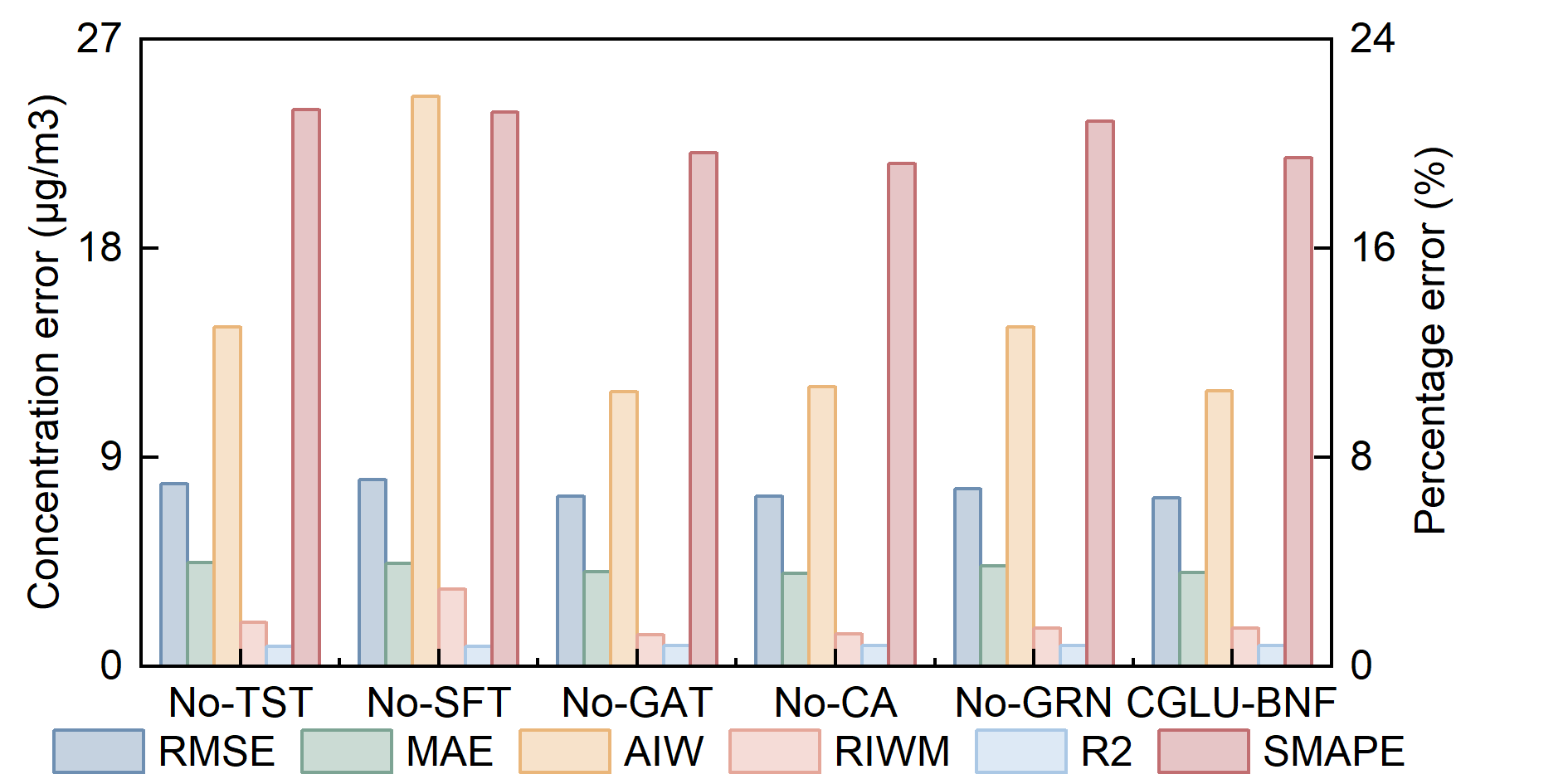}
	\caption{Contribution of different exogenous covariates to the $PM_{10}$ concentration prediction task}
	\label{fig8}
\end{figure}

To quantify the marginal value of exogenous covariates for $PM_{10}$ forecasting, we adopt a single variable incremental protocol on the Hong Kong dataset while holding the model architecture and training procedure fixed. In each run, we retain only the spatio-temporal features and add one exogenous variable to assess its contribution. As shown in Fig.\ref{fig8}, $PM_{2.5}$ delivers the largest accuracy gain. Gaseous pollutants yield smaller but meaningful improvements, with $NO_{2}$ contributing the most among them.

This phenomenon has a reasonable physical basis. $PM_{2.5}$ and $PM_{10}$ co-vary because they share emission sources and undergo coupled aerosol-mass evolution. Including $PM_{2.5}$ therefore improves accuracy and narrows prediction intervals. Among the gases, $NO_{2}$ is the strongest predictor of $PM_{10}$ because it serves as a robust proxy for traffic-related emissions.

\section{Conclusion and Future Works}
\label{conclusion}
This study introduces CGLU-BNF, a Bayesian deep learning framework for air quality prediction, with three key advantages: (i) It eliminates the need for preprocessing steps such as spatial interpolation or temporal padding, enabling direct extraction of spatio-temporal feature evolution from incomplete observations while simultaneously quantifying predictive uncertainty. (ii) Its feature encoding module, which integrates Fourier functions with a graph attention mechanism, effectively captures multi-scale spatial dependencies and seasonal temporal patterns across different frequencies. (iii) Its paired multiple channel gated learning unit adaptively filters and amplifies informative features, substantially improving predictive accuracy for sparse datasets.

Experimental results demonstrate that the proposed model substantially outperforms other air quality prediction methods with uncertainty estimation across four common missing data scenarios: random missing, node missing, timestamp missing, and spatio-temporal block missing. Its robustness is further validated in varying prediction horizon tasks, where it consistently surpasses the state-of-the-art BayesNF. Ablation studies also confirm the effectiveness of the individual strategies and modules within CGLU-BNF.

Future work will focus on architectural optimizations to accelerate training for long horizon forecasting on large scale datasets. We will also examine model performance under extremely sparse observation regimes, such as in-vehicle mobile monitoring.

\section{CRediT Authorship Contribution Statement}
The model was designed and implemented by Yuzhuang Pian and Taiyu Wang. Evaluations were designed by Yuzhuang Pian. Implemented by Yuzhuang Pian, Rui Xu and Shiqi Zhang. Yonghong Liu provided guidance and oversight. Yuzhuang Pian drafted the manuscript, all authors contributed to its revision and completion.

\section{Declaration of Competing Interest}
The authors declare that they have no known competing financial interests or personal relationships that could have appeared to influence the work reported in this paper.

\section{Acknowledgments}
This work was supported by the Fundamental Research Key Program of Shenzhen (No. JCYJ20241202130016022) and the Guangzhou National Games Air Quality Enhancement Project (No. SYSU-76160-20240710-0002).

\bibliographystyle{IEEEtran}
\bibliography{ref_abb}

\vfill

\end{document}